\documentclass[12pt,onecolumn,journal,transmag]{IEEEtran}

\usepackage[noadjust]{cite}

\usepackage{graphicx}
\usepackage{amsfonts}
\usepackage{epstopdf,gensymb}
\usepackage[cmex10]{amsmath}
\usepackage{amsmath}
\usepackage{amssymb}
\usepackage{dsfont}

\usepackage[tight,footnotesize]{subfigure}
\usepackage{caption}[2007/12/23]
\usepackage{color}
\usepackage{multirow}
\usepackage{booktabs}
\title{Beamforming through regularized inverse problems in ultrasound medical imaging}

\author{Teodora Szasz, Adrian Basarab, and Denis Kouam\'e}

\begin{document}
\maketitle
\bstctlcite{IEEE:BSTcontrol}

\begin{abstract}
Beamforming in ultrasound imaging has significant impact on the quality of the final image, controlling its resolution and contrast. Despite its low spatial resolution and contrast, delay-and-sum is still extensively used nowadays in clinical applications, due to its real-time capabilities. The most common alternatives are minimum variance method and its variants, which overcome the drawbacks of delay-and-sum, at the cost of higher computational complexity that limits its utilization in real-time applications. 

In this paper, we propose to perform beamforming in ultrasound imaging through a regularized inverse problem based on a linear model relating the reflected echoes to the signal to be recovered. Our approach presents two major advantages: i) its flexibility in the choice of statistical assumptions on the signal to be beamformed (Laplacian and Gaussian statistics are tested herein) and ii) its robustness to a reduced number of pulse emissions. The proposed framework is flexible and allows for choosing the right trade-off between noise suppression and sharpness of the resulted image. We illustrate the performance of our approach on both simulated and experimental data, with \textit{in vivo} examples of carotid and thyroid. Compared to delay-and-sum, minimimum variance and two other recently published beamforming techniques, our method offers better spatial resolution, respectively contrast, when using Laplacian and Gaussian priors.
                
\end{abstract}

\begin{IEEEkeywords}
Adaptive beamforming, linear inverse problems, beamspace processing, Basis Pursuit, Least Squares
\end{IEEEkeywords}

\section{Introduction}

\IEEEPARstart{U}{ltrasound} (US) imaging is one of the most fast-developing medical imaging techniques, allowing non-invasive and ultra-high frame rate procedures at reduced costs. Cardiac, abdominal, fetal, and breast imaging are some of the applications where it is extensively used as diagnostic tool. The new advances in beam formation, signal processing, and image display enlarge the US imaging potential to other fields like brain surgery, or skin imaging (e.g. \cite{mace_functional_2011} and \cite{wortsman_real-time_2004}).

In a classical US scanning process, short acoustic pulses are transmitted through the region-of-interest (ROI) of the human body. The backscattered echo signals, also called raw radiofrequency (RF) data, are then processed for creating RF beamformed lines. Beamforming (BF) plays a key role in US image formation, influencing the resolution and the contrast of final image. The most used BF method is the standard delay-and-sum (DAS) which consists in delaying and weighting the reflected echoes before averaging them. So far, its simplicity and real-time capabilities make it extensively used in ultrasound scanners. However, its weights are independent on the echo signals, resulting in beamformed signals with a wide mainlobe width and high side-lobe level. Consequently, the resolution and the contrast of final image are relatively low \cite{jensen_synthetic_2006}. Several adaptive beamformers (with weights dependent on data) from array processing literature were applied to US, the most common being the Capon or minimum variance (MV) BF \cite{viola_adaptive_2005}. It offers a very good interference rejection and better resolution than DAS, allowing higher contrast \cite{rindal_understanding_2014}. However, this method uses an estimated covariance matrix of the data and its main issue is the high computation complexity due to the calculation of the inverse covariance matrix. To overcome this, many improved versions of MV have been recently proposed (e.g. \cite{chen_improved_2013} and \cite{kim_fast_2014}), but still not adequate for real-time applications. In practice, in order to provide well-conditioned covariance matrices, diagonal loading, time and spatial averaging approaches were investigated, see e.g. \cite{synnevag_adaptive_2007}, respectively \cite{asl_minimum_2009}. 

Recently, to improve the MV BF, Nilsen \textit{et. al.} proposed a beamspace adaptive beamformer, BS-Capon, and unlike MV BF, they based their BF method on orthogonal beams formed in different directions \cite{nilsen_beamspace_2009}. This technique was also applied by Jensen \textit{et al.} to develop an adaptive beamformer based on multibeam covariance matrices \cite{jensen_approach_2012}, called multi-beam Capon beamformer. In their works, a covariance matrix is calculated for each range in the image, based on the idea that the beams were transmitted with different angles. Thus, the authors of \cite{jensen_approach_2012} were able to reduce the computation time of MV BF, while improving the resolution of the point-like reflectors. Following a similar idea, Jensen and Austeng \cite{jensen_iterative_2014} applied to US imaging a method proposed initially by Yardibi \textit{et al.} \cite{yardibi_nonparametric_2008}, called the iterative adaptive approach (IAA). They obtained better defined cyst-like structures compared to conventional DAS, and better rendering than MV.  

The work presented here uses a similar idea of beamforming range by range. However, inspired from the source localization problems, we represent, for each range, the BF as a linear direct model relating the raw samples to the desired lateral profile of the RF image to be beamformed. This formalism allows us to invert the problem by imposing standard regularizations such as the $\ell_{1}$ or $\ell_{2}$-norms. These choices are motivated by the existing works in US image enhancement, that are typically based on Laplacian (e.g., \cite{michailovich_blind_2007}) or Gaussian (e.g., \cite{jirik_two-dimensional_2008}) priors. Thus, the major contribution of this paper is the improvement of the existing beamforming techniques by combining the proposed direct model formulated in the lateral direction of the images with a regularized inversion approach. Moreover, we incorporated the proposed method with a beamspace processing technique, in order to highly reduce the number of the required US emissions. 

In contrast to existing BF methods in US imaging using regularized inverse problem approaches (see e.g., \cite{lavarello_regularized_2006, lingvall_method_2004, lingvall_time-domain_2007, madore_reconstruction_2012, viola_time-domain_2008, wan_post-beamforming_2009}), our method does not use the system point spread function (PSF) in the direct model or in the inversion process. Thus, the proposed BF technique does not require any experimental measurement (e.g., \cite{ellis_super-resolution_2010}) or estimation of the PSF (e.g., \cite{jensen_deconvolution_1992}) \cite{michailovich_blind_2007}, \cite{zhao_joint_2015}). 

Laplacian and Gaussian statistics, two of the most common regularizations in such imaging problems (including US imaging applications such as deconvolution), are considered herein, allowing the reader to observe their influence on the results. Furthermore, our method opens new tracks for more complex regularization terms (e.g. \cite{zou_regularization_2005, ChBaKo2016.1, Michailovich2015}) to further improve the results. The proposed approaches, generically named Basis Pursuit beamforming (BP BF), respectively Least Squares beamforming (LS BF) in this paper, were evaluated using different Field II simulated data and \textit{in vivo} carotid and thyroid experimental data. Finally we compared our BF techniques with four existing beamformers: the conventional DAS, MV, multi-beam Capon, and IAA.

The reminder of the paper is organized as follows. First, in Section \ref{sec:bkgd_us} we summarize the theoretical background of BF. In Section \ref{sec:BF_inverse} we describe the proposed BF method from a regularized inverse problem perspective. Details about the experiments and the results are given in Section \ref{sec:resdisc}, and Section \ref{sec:conclusion} concludes the paper.

\section{Background} \label{sec:bkgd_us}

\begin{figure}
  \centering
  \includegraphics[width=0.45\textwidth]{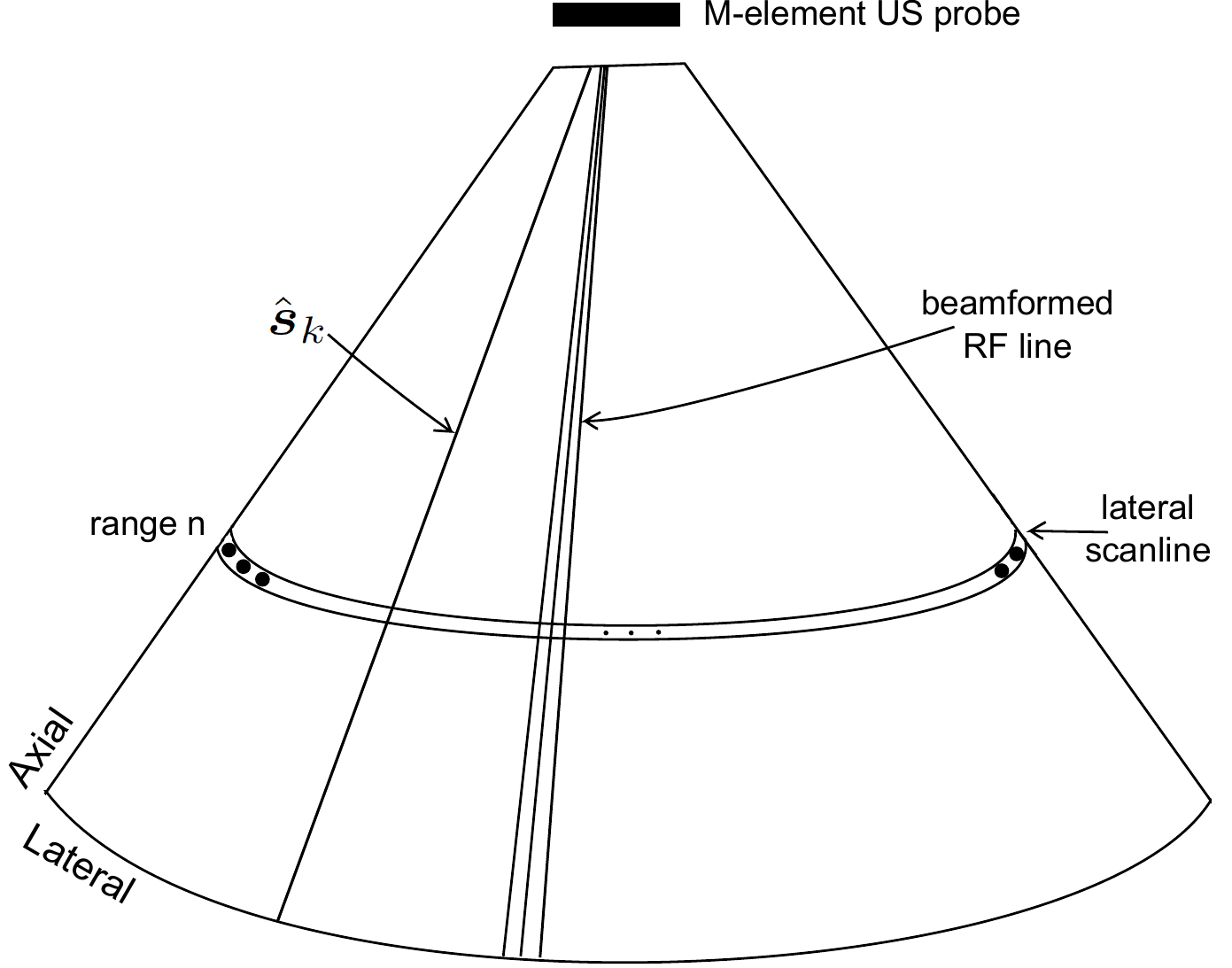}
  \caption{The main US imaging elements used to adapt the array processing beamforming methods discussed in Section~\ref{sec:bkgd_us} and Section~\ref{sec:BF_inverse} to medical US imaging.}
\label{fig:setup}
\end{figure}

The main elements used to model the BF process are depicted in Fig.~\ref{fig:setup}. We consider, without loss of generality, the particular setup of an $M$-element US probe ($M$ can be also the number of active elements of the probe), with the transducer's elements denoted by $u_{m}$, with $m=0,\cdots M-1$. We consider a trivial change of variable, such that the position of the $m$-th element is:
\begin{equation}
p_{m}=[m-(M-1)/2](\lambda/2), \ \ \ m=0,1, \cdots, M-1,
\label{eq:position_elem}
\end{equation}
with the probe's elements positioned symmetrically around origin. We have considered, as example, the pitch equal to $\lambda/2$ (the spacing between elements is half of the wavelength $\lambda=c/f_{0}$, $c$ denoting the speed of the sound through soft tissue, and $f_{0}$ the transducer's center frequency). 

A series of $K$ focused beams are transmitted with different incident angles $\theta_k$, $ k=1,\cdots K$. The returning echoes are recorded using the same US probe, being time-delayed, such that the time-of-flight is compensated, so the backscatter from the point of interest is summed up coherently. If we consider that each of the recorded raw signal after the time-delay compensation has N time samples, the size of the recorded data from all the $K$ directions will be $M\times N \times K$. Finally, the final RF US image is a collection of RF beamformed lines, each of which being the result of beamforming the raw RF signals coming from an emission in the direction $\theta_{k}$, $k\in \{1,\hdots,K\}$, using $M$ elements of the tranducer.

The classical DAS BF can be expressed as:
\begin{equation}
\begin{split}
\hat{\boldsymbol{s}}_{k}(n) = \sum_{m=1}^{M}w_{m}\boldsymbol{y}_{m}^{(k)}(n-\Delta_{m}(n)), \ \ \ &n=1,\cdots ,N,\\
&k=1,\cdots,K,
\end{split}
\label{eq:BF_DAS}
\end{equation}
where $\Delta_{m}(n)$ is the time delay for focusing at the point of interest sample, being dependent on the distance between the $m$-th element and the point of interest, $w_{m}$ are the BF weights, $\boldsymbol{y}_{m}^{(k)} \in \mathbb{C}^{N\times 1}$ is the raw data received by the $m$-th element of the US probe, corresponding to the emission steered at angle $\theta_k$. A simplified form of \eqref{eq:BF_DAS} can be formulated as: 

\begin{equation}
\hat{\boldsymbol{s}}_{k}=\boldsymbol{w}^{H}\boldsymbol{y}_{k},
\label{eq:BF_DAS_2}
\end{equation}
where $\boldsymbol{y}_{k} \in \mathbb{C}^{M \times N} $ is the time-compensated version of $\boldsymbol{y}_{m}^{(k)}$ in \eqref{eq:BF_DAS} for the $k$-th emission (for the sake of generality, we consider $\boldsymbol{y}_{k}$ to be complex-valued data), $\boldsymbol{w}$ is the vector of the beamformer weights of size $M \times 1$, and $(\cdot)^{H}$ represents the conjugate transpose. 
DAS BF selects the weights independent on data, solving:
\begin{equation}
\min_{\boldsymbol{w}}\boldsymbol{w}^{H}\boldsymbol{w}, \ \ \ \  \text{such that} \ \ \ \ \boldsymbol{w}^{H}\boldsymbol{1}=1, 
\label{eq:das_crit}
\end{equation}
where $\boldsymbol{1}$ is a length $M$ column-vector of ones since the raw data was focused using time-delays. The solution of \eqref{eq:das_crit} is: 
\begin{equation}
\boldsymbol{w}_{DAS}=\frac{\boldsymbol{1}}{M}.
\label{eq:das_w}
\end{equation}
If we replace \eqref{eq:das_w} in \eqref{eq:BF_DAS_2} we get:
\begin{equation}
\hat{\boldsymbol{s}}_{k}=\frac{1}{M} \boldsymbol{1}^T \boldsymbol{y}_{k},
\label{eq:BF_DAS_fin}
\end{equation}
where $\{\cdot\}^{T}$ denotes the transpose.
A common technique used in US is to apply weighting functions such as Hanning, or Hamming apodizations to \eqref{eq:BF_DAS_fin} to further reduce the sidelobes of $\hat{\boldsymbol{s}}_{k}$, resulting in improved contrast of the beamformed image, at the cost of a slight lateral spatial degradation.

Further details about adaptive BF in US imaging (i.e., the methods used for comparision purpose) and about beamspace processing are provided in Appendices \ref{subsec:mv}, \ref{subsec:multibean_caponBF}, \ref{subsec:iaa},  respectively in Appendix \ref{subsec:beamsp_beam_Butler}.   

\section{Proposed method: Beamforming through regularized inverse problems} \label{sec:BF_inverse}

\subsection{Model formulation}
\label{subsec:model}

\begin{figure}
  \centering
  \includegraphics[width=0.45\textwidth]{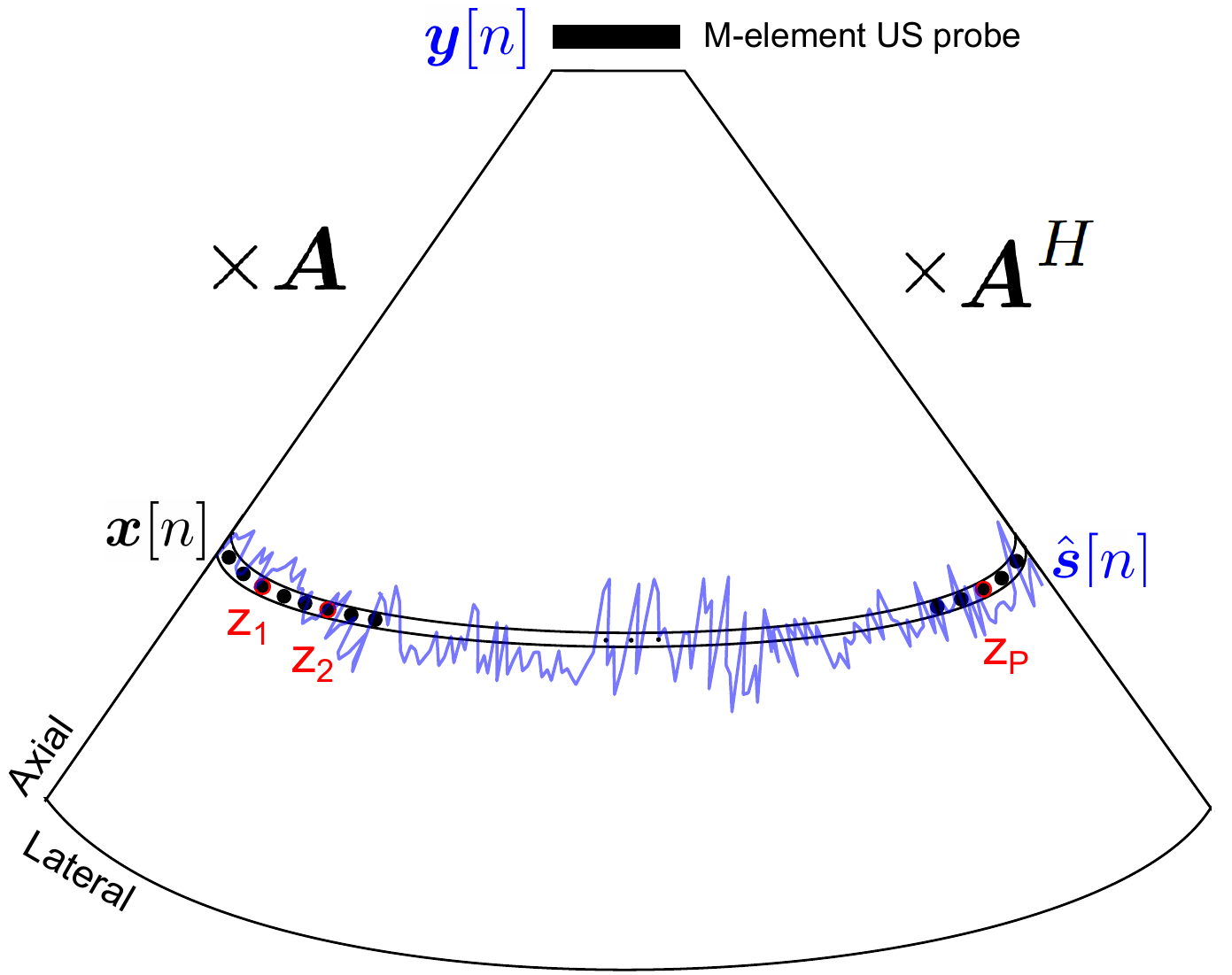}
  \caption{The elements used to form the proposed model.}
\label{fig:our_model}
\end{figure}

The main elements used to model the proposed method are depicted in the Fig.~\ref{fig:our_model}. For sake of simplicity, let us focus our problem at a time-sample (range) $n$. The proposed BF method is sequentially applied in the same manner to each range. If $\boldsymbol{y}_{k}[n] \in \mathbb{C}^{M \times 1}$ is the raw data after the compensation of the time-of-flight for the $k$-th steering direction $\theta_{k}$, we can form the steering vectors as in \eqref{eq:theta_k_iaa}. Let $\boldsymbol{A}$ be the $M \times K$ steering matrix containing the steering vectors in \eqref{eq:theta_k_iaa} for all $\theta_{k}$ directions, $k=1, \cdots ,K$:

\begin{equation}
\boldsymbol{A} = [\boldsymbol{a}_{\theta_{1}}, \boldsymbol{a}_{\theta_{2}}, \cdots ,\boldsymbol{a}_{\theta_{K}}]. 
\label{eq:A_matrix}
\end{equation}
Note that $\boldsymbol{A}$ is known and depends on the positions of the probe elements and on the locations to beamform. Thus, it is independent on the actual positions of the reflectors.

For each range $n$, we want to estimate the signal corresponding to a reflector as a function of its location, that will contain dominant peaks at reflector positions. Thus, the main difference from the multi-beam Capon beamforming method is that instead of calculating the values of the weights $\boldsymbol{w}_{\theta,n}$ as in \eqref{eq:mv_w_lat}, that are further used to calculate the reflector's signal, we are directly estimating the corresponding signal by considering the raw data $\boldsymbol{y}_{k}[n]$ as observations. In other words, we want to obtain an estimate of the reflected echo $\boldsymbol{x}[n] \in \mathbb{C}^{K \times 1}$ through the observations $\boldsymbol{y}_{k}[n]$. Unfortunately one difficulty arises: since $\boldsymbol{y}_{k}[n]$ is corresponding to only one emission, modeling our problem using raw data as observations to estimate the reflectors, requires high computational cost, since we are dealing with multiple directions. We recall that the size of raw data in our problem at a range $n$, is $M \times K$. To overcome this issue, motivated by the results in \cite{malioutov_sparse_2005} and \cite{fuchs_linear_1996}, we propose to use the DAS beamformed data instead of the original raw data. In addition to data dimensionality reduction, it was shown in \cite{malioutov_sparse_2005} and \cite{fuchs_linear_1996} that more accurate results may be achieved by proceeding in this way. Thus, we can formulate our model as follows:

\begin{equation}
\hat{\boldsymbol{s}}[n] = (\boldsymbol{A}^{H}\boldsymbol{A})\boldsymbol{x}[n]+\boldsymbol{g}[n],
\label{eq:model_das_not_beamspaced}
\end{equation}
where $\hat{\boldsymbol{s}}[n] \in \mathbb{C}^{K \times 1}$ is a lateral scanline of the DAS beamformed image formed as discussed in the Section \ref{sec:bkgd_us}, $\boldsymbol{A}$ is the steering matrix formed with \eqref{eq:A_matrix}, and $\boldsymbol{g}[n]$ an additive white Gaussian noise. If we denote by $\hat{\boldsymbol{S}}$ the DAS beamformed image of size $K\times N$, formed by juxtaposing the DAS RF lines $\hat{\boldsymbol{s}}_{k}$ expressed in \eqref{eq:BF_DAS_fin} for all $K$ directions, we consider $\hat{\boldsymbol{s}}[n]$ the lateral scanline from $\hat{\boldsymbol{S}}$ taken at the time-sample (range) $n$. Note that, since the transducers are emitting the same pulse, we assumed that $\boldsymbol{x}[n]$ which is the unknown signal, is the same for all $K$ emissions, and for all transmitters (see e.g. \cite{du_review_2008}).

The role of the multiplication of the steering matrix $\boldsymbol{A}$ with its conjugate transpose $\boldsymbol{A}^{H}$ in \eqref{eq:model_das_not_beamspaced} is to relate the position of the elements with the position of all $K$ reflectors on a scanline. This relation is a result of considering on the one hand that the elements are impinging to the reflectors situated on the lateral scanline (the multiplication of $\boldsymbol{A}$ with $\boldsymbol{x}[n]$), while on the other hand the reflectors are impinging to the elements through their reflected pulses (the multiplication with $\boldsymbol{A}^{H}$). Hence, the result of the DAS beamformed scanline $\hat{\boldsymbol{s}}[n]$ is related to the unknown signal $\boldsymbol{x}[n]$ through a direct linear model. Fig.~\ref{fig:our_model} offers a schematic representation of our model in \eqref{eq:model_das_not_beamspaced}. Thus, after the compensation of the time-of-flight, the received raw data, $\boldsymbol{y}[n]$ at a range $n$ is formed by multiplying the steering matrix $\boldsymbol{A}$ with the desired signal $\boldsymbol{x}[n]$, $\boldsymbol{y}[n]=\boldsymbol{A} \boldsymbol{x}[n]$. This multiplication could be sufficient for describing the proposed model if we are considering the raw data $\boldsymbol{y}[n]$, as observations. However, since we are using the DAS beamformed data instead of the original raw data, we further take into account the geometrical relationship between the potential sources and the elements of the probe (through the multiplication with $\boldsymbol{A}^{H}$).

\subsection{Beamspace processing}
\label{subsec:beamspace_us}

In order to solve $\eqref{eq:model_das_not_beamspaced}$, we firstly apply beamspace processing, a common tool used in source localization approaches that reduces the computational complexity, while improving the resolution, and reducing the sensitivity to the position of the sensor (see e.g. \cite{fuchs_linear_1996} and \cite{tian_beamspace_2001}). Its main purpose in US is to reduce the number of the US emissions, thus reducing the acquisition time and the computational load required by the BF process. We should note that our method of transforming the data into beamspace domain is totally different from the technique resumed in Appendix \ref{subsec:beamsp_beam_Butler}. The main reason is that, by using the beamspace processing presented in \cite{nilsen_beamspace_2009}, we need all the acquired raw data for applying beamspace processing as described in \eqref{eq:beamspace_transf}. Hence, even if on one hand, the computational complexity required by the estimated covariance matrix inversion is reduced, on the other hand, it is increased by the operations required to transform the entire set of the raw data into its beamspaced correspondents.

To overcome this, we based our idea on the beamspace processing techniques proposed in array processing (notably in source localization). More specifically, Malioutov \textit{et al.} \cite{malioutov_sparse_2005} used a method that maps the data from full dimension space of the directions (DS) into a lower dimension beamspace (BS) through a linear transform prior to source localization processing. In our case, for each range $n$ we project the data resulted by applying DAS BF, $\hat{\boldsymbol{s}}[n]$, in BS before beamforming it through regularized inverse problems. To emphasize, $\hat{\boldsymbol{s}}[n]\in \mathbb{C}^{K \times 1}$ is projected on a lateral sampled grid of $P<<K$ locations. In other words, the proposed BF method, contrarily to all the other discussed BF methods, uses only $P$ focused emitted beams among all the $K$ transmissions to beamform a particular lateral scanline of $K$ samples. Thus, the number of emissions is reduced by a factor of $\frac{K}{P}$. This will result in a reduced dimensionality of the data compared with the other BF methods, and an improved computational complexity compared with MV, multi-beam Capon, and IAA.

Let $\boldsymbol{z}[n]\in \mathbb{C}^{P \times 1}$ be the beamspace transformed vector formed by sampling the DAS beamformed lateral scanline $\hat{\boldsymbol{s}}[n]$ on a grid of $P$ locations, see Fig.~\ref{fig:our_model}:

\begin{equation}
\boldsymbol{z}[n] = \boldsymbol{D}^{H}\hat{\boldsymbol{s}}[n], 
\label{eq:s, beamspaced}
\end{equation}
where $\boldsymbol{D}$ of size $K \times P$ is the beamspace decimation matrix, that will reduce the dimensionality of a vector from $K \times1$ to $P \times1$. Hence, since the decimation factor is $\frac{K}{P}$, $\boldsymbol{D}$ has all elements zero, except the elements $d_{i,j}$ with $j=\frac{K}{P}i$, that will get the value $1$. Similarly, the beamspaced steering matrix $\boldsymbol{A}_{BS}^{H}$ of size $P \times M$ is formed, composed of the $P$ beamspaced steering vectors:

\begin{equation}
\boldsymbol{A}_{BS}^{H}=\boldsymbol{D}^{H}\boldsymbol{A}^{H}
\label{eq:a_beamspaced}
\end{equation}

Concretely, we form $\boldsymbol{A}_{BS}\in \mathbb{C}^{M \times P}$ by taking from $\boldsymbol{A}^{H}$ each $\frac{K}{P}$-th steering vector. So, the model formed by \eqref{eq:model_das_not_beamspaced} after applying beamspace processing, becomes:

\begin{equation}
\boldsymbol{z}[n] = \boldsymbol{D}^{H} \hat{\boldsymbol{s}}[n] = (\boldsymbol{A}_{BS}^{H}\boldsymbol{A})\boldsymbol{x}[n] + \boldsymbol{D}^{H}\boldsymbol{g}[n],
\label{eq:our_BS}
\end{equation}
where $\boldsymbol{x}[n]$ of size $K \times 1$ is the lateral profile at range $n$ to be estimated. Thus, we can see \eqref{eq:our_BS} as an inverse problem, where $\boldsymbol{z}[n]$ is the DAS beamformed data corresponding to $P < K$ emissions, and considered as the observation data.    

\subsection{Beamforming through regularized inverse problems} \label{sec:lp_bf}

Given the ill-posedness of the direct model in \eqref{eq:our_BS}, we propose hereafter to invert it using standard regularization techniques. For achieving this, a cost function, denoted by $J(\boldsymbol{x}[n])$, consisting into the linear combination of two terms is considered. The first term, denoted by $J_{1}(\boldsymbol{x}[n])$, represents the data attachment, while the second, denoted by $J_{2}(\boldsymbol{x}[n])$, is the regularization prior:

\begin{equation}
J(\boldsymbol{x}[n]) = J_{1}(\boldsymbol{x}[n]) + \lambda J_{2}(\boldsymbol{x}[n]),
\label{eq:cost_fct}
\end{equation}
where $\lambda$ is a scalar, called regularization parameter, that adjusts the trade-off between the fidelity to the data and the regularization term. A large $\lambda$ will strongly favor the \textit{a priori} about $\boldsymbol{x}[n]$, while a small $\lambda$ gives a high confidence to the observations. Keeping in mind that the additional noise in \eqref{eq:our_BS} is Gaussian, the data attachment term is expressed by an $\ell_{2}$-norm, giving the following cost function:
\begin{equation}
J(\boldsymbol{x}[n]) = ||\boldsymbol{z}[n] - (\boldsymbol{A}_{BS}^{H}\boldsymbol{A})\boldsymbol{x}[n]||_{2}^{2} + \lambda J_{2}(\boldsymbol{x}[n]).  
\label{eq:cost_fct_r}
\end{equation}

In this work, the choice of the regularization term $J_{2}(\boldsymbol{x}[n])$ is guided by the existing literature on statistical modelling of US images, previously applied to various applications such as image deconvolution or segmentation (e.g. \cite{michailovich_blind_2007}, \cite{yu_blind_2012}). It has thus been shown that Laplacian and Gaussian statistics are well adapted to model US images. For this reason, we give in the two following paragraphs the mathematical derivations and beamforming results using $\ell_{1}$ norm (the sum of absolute difference) and $\ell_{2}$ norm (or the Euclidean norm, that is the sum of squared difference) regularization terms. We should note that while the first will promote sparse solutions, the latter will promote smoother results.

The choice of a quadratic data fidelity term is related to the additive zero-mean Gaussian assumption on the noise. We emphasize that the noise considered in our paper is different from the multiplicative speckle noise generally assumed to affect envelope images in ultrasound imaging. Instead, the additive noise considered in our paper affects the RF data and is caused by the acquisition process. The same model has been previously used by several authors (e.g. \cite{michailovich_blind_2007} and \cite{du_review_2008}).
\vspace{1em}

\subsubsection{\textbf{Laplacian statistics through Basis Pursuit (BP)}} \label{sec:l1}

Considering that the signal $\boldsymbol{x}[n]$ to be beamformed follows Laplacian statistics, the minimization of the cost function in \eqref{eq:cost_fct_r} turns into the optimzation procedure in \eqref{eq:cost_fct_fin}, usually called Basis Pursuit (BP) in the literature \cite{chen_atomic_1998}.

\begin{equation}
\boldsymbol{x}_{BP}[n]=\mathop{\rm argmin}\limits_{\boldsymbol{x}[n]}(||\boldsymbol{z}[n] - (\boldsymbol{A}_{BS}^{H}\boldsymbol{A})\boldsymbol{x}[n]||_{2}^{2} + \lambda ||\boldsymbol{x}[n]||_{1}),  
\label{eq:cost_fct_fin}
\end{equation}
where $||\cdot||_{1}$ denotes the $l_{1}$-norm.
The minimization problem \eqref{eq:cost_fct_fin} is convex, hence continuous, and has one global minima for any $\lambda > 0$.

Herein, we used the well known YALL1 to solve \eqref{eq:cost_fct_fin} \cite{_download_????}, a software package that contains implementation of alternating direction method (ADM), that solves also BP. A comparison of the six most used BP implementations is done in \cite{lorenz_solving_2015} and three of them were also compared in \cite{huang_comparison_2010} with application in underwater acoustics, where is shown that YALL1 gives best performances for real-time applications.

\vspace{1em}
   
\subsubsection{\textbf{Gaussian statistics through Least Squares (LS)}} \label{sec:l2}

To achieve smooth solutions of the proposed BF method, we modeled our problem with an $\ell_{2}$-norm based minimization function, and we used the Tikhonov regularized least-squared method (or rigid regression) for solving it \cite{a_solution_1963}. The cost function is of the form:

\begin{equation}
\boldsymbol{x}_{LS}[n]=\mathop{\rm argmin}\limits_{\boldsymbol{x}[n]}(||\boldsymbol{z}[n] - (\boldsymbol{A}_{BS}^{H}\boldsymbol{A})\boldsymbol{x}[n]||_{2}^{2} + \lambda ||\boldsymbol{x}[n]||_{2}^{2}), 
\label{eq:cost_fct_ls}
\end{equation}
where $||\cdot||_{2}$ denotes the $\ell_{2}$-norm. For solving \eqref{eq:cost_fct_ls} we used its analytical solution:

\begin{equation}
\boldsymbol{x}_{LS}[n] = ((\boldsymbol{A}_{BS}^{H}\boldsymbol{A})^{H}(\boldsymbol{A}_{BS}^{H}\boldsymbol{A}) + \lambda \boldsymbol{I}_{K \times K})^{-1}(\boldsymbol{A}_{BS}^{H}\boldsymbol{A})^{H}\boldsymbol{z}[n],  
\label{eq:diff_aprox}
\end{equation}
where $\boldsymbol{I}_{K \times K}$ denotes the identity matrix of size $K \times K$. 

In order to obtain the BP and LS beamformed images, for each time sample $n$, with $n\in\{1,\hdots,N\}$, we estimate its corresponding lateral scanline $\boldsymbol{x}_{BP}[n]$ (using BP BF method) or $\boldsymbol{x}_{LS}[n]$ (using LS BF method), and we are juxtaposing all the obtained scanlines, in the axial direction of the image. 

\section{Experiments} \label{sec:experiments}

\begin{table*}
\caption{Parameters of simulated and experimental images}

\begin{center}
\begin{tabular}{ l l l l l l}
\hline
\noalign{\vskip 2mm}
\textbf{Parameters for simulation of}: & Point scatterers & Cyst simulation & Simulation of cardiac Image & Experimental carotid & \textit{In vivo} thyroid \\
 & \textbf{Fig.~\ref{fig:comp_reflectors}}& \textbf{Fig.~\ref{fig:comp_hypo_phantom}} & \textbf{Fig.~\ref{fig:comp_cardio}} & \textbf{Fig.~\ref{fig:comp_carotid}} & \textbf{Fig.~\ref{fig:comp_thyroid}} \\
\noalign{\vskip 2mm}
\hline
\noalign{\vskip 2mm}
\multicolumn{6}{c}{\textbf{Transducer}} \\[1ex]
\hline
\noalign{\vskip 2mm}
Transducer type & \multicolumn{5}{c}{Linear array}  \\
Transducer element pitch [$\mu$m] & 256 & 256 & 192.5 & 110 &120 \\
Transducer element kerf [$\mu$m] & 20 & 20 & 38.5 & 25 \\
Transducer element height [mm] & 5  & 5  & 14  & 4  \\
Central frequency, $f_{0}$ [MHz] & 3 & 3 & 4 & 7 & 7.2 \\
Sampling frequency, $f_{s}$ [MHz] & 100 & 100 & 40 & 40 & 40 \\
Speed of sound, $c$ [m/s] & \multicolumn{5}{c}{1540}  \\
Wavelength [$\mu$m] & 513.3 & 513.3 & 385 & 220 \\
Excitation pulse & \multicolumn{5}{c}{Two-cycle sinusoidal at $f_{0}$}\\[1ex]
\hline
\noalign{\vskip 2mm}
\multicolumn{6}{c}{\textbf{Synthetic Aperture Emission}} \\[1ex]
\hline
\noalign{\vskip 2mm}
Receive Apodization & \multicolumn{4}{c}{Hanning} \\
Number of transmitting elements & 64 & 64 & 64 & 128 & 128 \\
Number of receiving elements & 64 & 64 & 64 & 128 & 128 \\
Number of emissions ($K$) & 260 & 260 & 204 & 192 & 312 \\[1ex]
\hline
\noalign{\vskip 2mm}
\multicolumn{6}{c}{\textbf{The values of $\lambda$ for the simulated and experimental images}} \\[1ex]
\hline
\noalign{\vskip 2mm}
BP & 0.5 & 0.5 & 0.2 & 0.5 & 5 \\
LS & 0.7 & 1 & 0.5 & 1 & 1 \\
\hline
\end{tabular}
\end{center}
\label{tab:sim_param}
\end{table*} 

For evaluating the proposed BP BF and LS BF approaches, we considered different types of simulated and experimental data. We compared our BF results with DAS (Section \ref{sec:bkgd_us}), MV (Appendix \ref{subsec:mv}), multi-beam Capon (Appendix \ref{subsec:multibean_caponBF}), and IAA (Appendix \ref{subsec:iaa}) BF methods. The simulations were made using the Field II program (see e.g. \cite{jensen_calculation_1992} and \cite{jensen_simulation_2004}). The first simulation include a sparse medium, the second one contains a circular hypoechoic cyst in a medium with speckle, and the third one contains a simulation of the Short Axis (SAx) view of a cardiac image, as suggested in \cite{alessandrini_simulation_2012}. The first experimental data consists in a carotid that was recorded with an Ultrasonix MDP research platform. Finally, the second experimental data contains a thyroid medium with a malignant tumor. The thyroid data was recorded with the Sonoline Elegra clinical scanner, modified for research purpose. The parameters of the simulated and experimental data are presented in the Table~\ref{tab:sim_param}. Note that for MV and multi-beam Capon beamformers, spatial and temporal averaging, as well as diagonal loading technique are used for the estimation of the covariance matrix, as discussed in the Section \ref{subsec:mv}. 

An important aspect is that, when applying the proposed BF methods, five times ($\frac{K}{P}=5)$ less emissions were used in the beamforming process, by applying the beamspace processing presented in the Section \ref{subsec:beamspace_us}. This hangs on for all the examples we are presenting in this paper. For these examples, reducing five times the US transmissions is optimal in terms of gain in resolution, while reducing computational time.

The values of the regularization parameter $\lambda$ for all the presented examples are grouped in the Table~\ref{tab:sim_param}. The optimal value of $\lambda$ was chosen manually. We emphasize that this was the case for all hyperparameters of all comparative methods. Several studies exist in the literature for automatic estimation of the regularization hyperparameter (e.g. \cite{chen_simulation_2015}, \cite{galatsanos_methods_1992}, and \cite{ramani_regularization_2012}) that can improve the robustness of the proposed methods, at increased computational cost. Nonetheless, their implementation is beyond the scope of this paper.

\subsection{Parameters for the comparative methods} \label{sec:param_comp}
The results of MV beamforming were obtained by using the implementation described in \cite{asl_minimum_2009}. The length of the spatial averaging window, $L$ was defined as half of the number of the probe's elements, i.e. $L=\frac{M}{8}$. A temporal window of $10$ samples was used in our examples and the diagonal loading parameter was fixed to $\Delta = \frac{1}{10L}$. The adaptive coherence method was applied to the MV BF method. 

The results of multibeam-Capon were obtained by using the multi-beam approach suggested in \cite{jensen_approach_2012}. The $K$ emissions were uniformly distributed between $\pm 30^{\circ}$. The beamspace transform down to 33 dimensions was applied, able to retain the variance for incoming narrowband far-field signals. The diagonal loading factor was set to $0.01$.

Finally, for IAA implementation we used the source code provided by the authors in \cite{jensen_iterative_2014}. The number of iterations was set to 15 for our examples.

Note that for all the comparative methods, several parameters need to be carefully tuned in order to obtain acceptable results. However, using the proposed approach, only the regularization hyperparameter $\lambda$ needs to be set.

\subsection{Simulated point reflectors} \label{sec:sim_point}

The medium contains 5 point reflectors, 4 of them aligned in pairs of 2 and separated by 4 mm, and the other laterally centered at 0 mm. They are located at axial depths ranging from 63 to 68 mm, with a transmit focus at 65 mm and a dynamic receive focalisation.  

\subsection{Simulated phantom data} \label{sec:sim_phantom}

To evaluate the accuracy, contrast and resolution of the aforementioned beamformers, a hypoechoic cyst of radius 5 mm, located at the depth 80 mm, in a speckle pattern. The speckle pattern contains 50000 randomly placed scatterers, with Gaussian distributed amplitudes. This example was inspired from the simulation of a synthetic kidney example included in Field II software. The attenuation was not taken into account.

\subsection{Simulated cardiac image} \label{sec:sim_cardiac}

The SAx view is the cross-sectional view of the heart and is a well exploited perspective in echocardiography, containing information about the left ventricle (LV) and right ventricle (RV). In our simulation we visualize the LV. The transmit focus point is set at 65 mm. The final image is ultra-realistic, the amplitudes being related to an \textit{in vivo} cardiac image \cite{alessandrini_simulation_2012}. The number of scatterers was sufficiently large to produce fully developed speckle.

\subsection{\textit{In vivo} data: carotid}
The carotid ultrasound is a common procedure used to detect strokes or the risk of strokes due to the narrowing of the carotid arteries. The data was acquired from a healthy subject, with the Ultrasonix MDP research platform, attached with the parallel channel acquisition system, SonixDAQ. The linear ultrasonic probe L14-W/60 Prosonic$^{\copyright}$ (Korea) of 128 elements was used.

\subsection{\textit{In vivo} data: thyroid}
The thyroid ultrasound is done to visualize the thyroid gland to detect possible tumors or deformations. Two sets of data were acquired: first one, from a subject with a tumor and the other one from a healthy subject. For both acquisitions we used the clinical Sonoline Elegra ultrasound system modified for research purposes, and a 7.5L40 P/N 5260281-L0850 linear array transducer of Siemens Medical Systems, having the characteristics described in Table~\ref{tab:sim_param}.

\subsection{Image quality measures}
Three image quality metrics were evaluated: the contrast-to-noise ratio (CNR), the signal-to-noise ratio (SNR), and the resolution gain (RG). They were computed based on the envelope-detected signals independent of image display range.

Based on two regions $R_1$ and $R_2$ belonging to two different structures, CNR is defined as \cite{garayoa_study_2013}:

\begin{equation}
CNR = \frac{|\mu_{R_1} - \mu_{R_2}|}{\sqrt{\sigma_{R_1}^2 + \sigma_{R_2}^2}},
\label{eq:CNR}
\end{equation}
where $\mu_{R_1}$ and $\mu_{R_2}$ are the mean values in the regions $R_1$, respectively $R_2$, and $\sigma_{R_1}$ and $\sigma_{R_2}$ are the standard deviations of intensities in $R_1$, respectively $R_2$.

The SNR is defined as the ratio between the mean value $\mu$ and the standard deviation $\sigma$ in homogeneous regions \cite{jensen_iterative_2014}:

\begin{equation}
SNR = \frac{\mu}{\sigma}
\label{eq:SNR}
\end{equation}

The RG is defined in \cite{taxt_two-dimensional_2001} as the ratio between the normalized autocorrelation function with values higher than 3 dB (computed for the DAS beamformed image in our case), over the normalized autocorrelation function (higher than 3 dB) of the images formed by using the other aforementioned BF methods (MV, multi-beam Capon, IAA, BP, and LS). Note that a value of RG $>1$ needs to be achieved for achieving a better resolution than DAS beamformer.

\section{Results and Discussion} \label{sec:resdisc}

\subsection{Individual point reflectors}

\begin{figure}
      \center
      \includegraphics[width=0.45\textwidth]{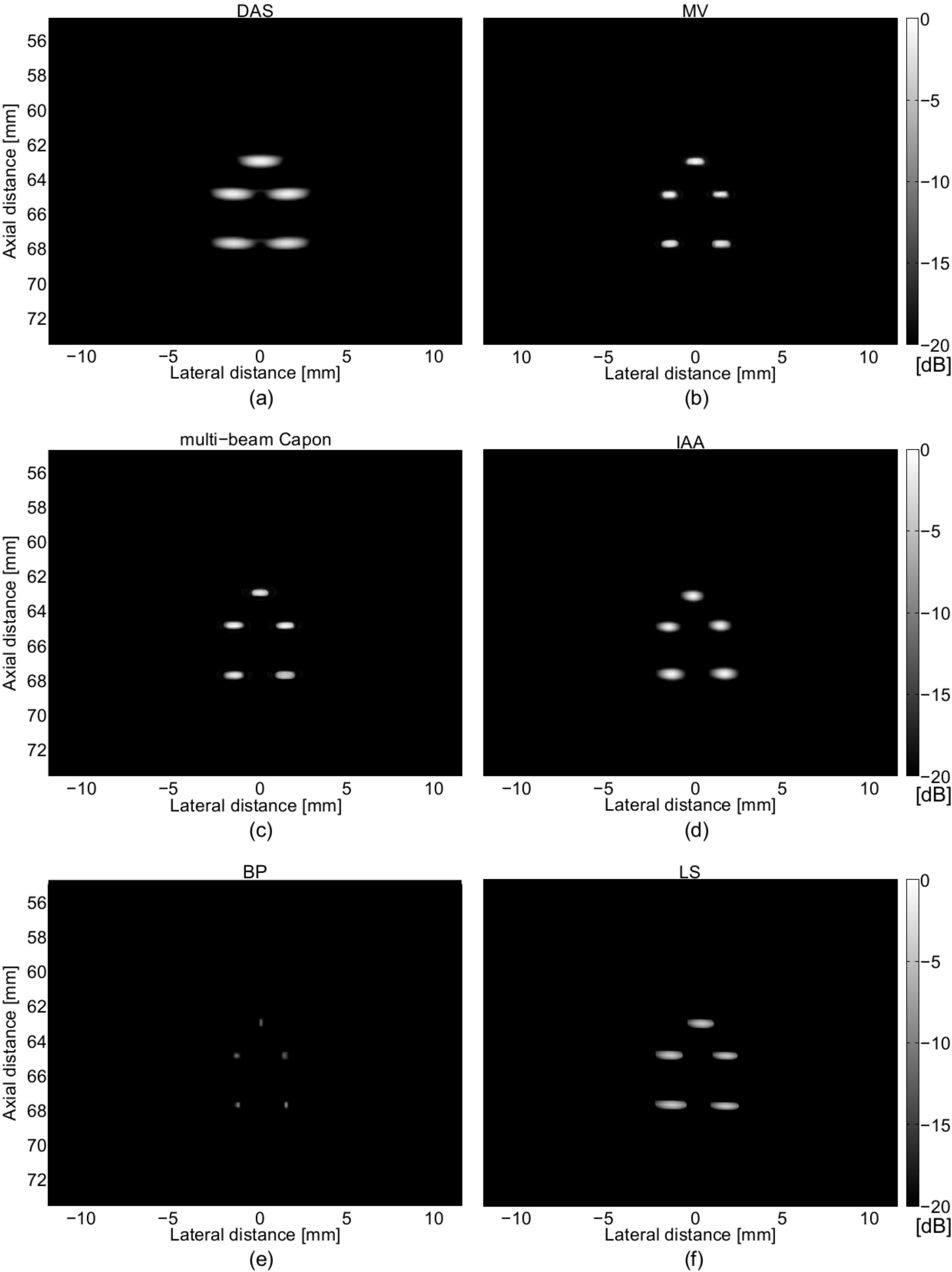}
      \hfil
      \caption{(a) DAS, (b) MV, (c) multi-beam Capon, (d) IAA, (e) BP, and (f) LS BF  results of the simulation of individual point scatterers.}
      \label{fig:comp_reflectors}
   \end{figure}
   
With this simulation we evaluate the potential of the proposed methods in sparse mediums. The resulted beamformed images are illustrated in the Fig.~\ref{fig:comp_reflectors}. The result of DAS BF is shown in Fig.~\ref{fig:comp_reflectors}(a). Using the MV BF, the lateral resolution is improved compared with DAS, IAA, and LS BF (see Fig.~\ref{fig:comp_reflectors}(b)), and it is comparable with the result of multi-beam Capon BF, Fig.~\ref{fig:comp_reflectors}(c). Concerning the IAA beamformed result, as stated in \cite{jensen_iterative_2014}, it gives better point-target resolvability than DAS, Fig.~\ref{fig:comp_reflectors}(d). The proposed BP BF have the best resolution of the point-like reflectors, being able to perfectly detect the 5 reflectors, by obtaining the most narrower mainlobes, due to the fact that BP results in a sparse representation of the beamformed signals, Fig.~\ref{fig:comp_reflectors}(e). As expected, LS beamformer results in solutions that tend to be smooth and regular, as in the Fig.~\ref{fig:comp_reflectors}(f).
   
\begin{figure}
      \center
      \includegraphics[width=0.45\textwidth]{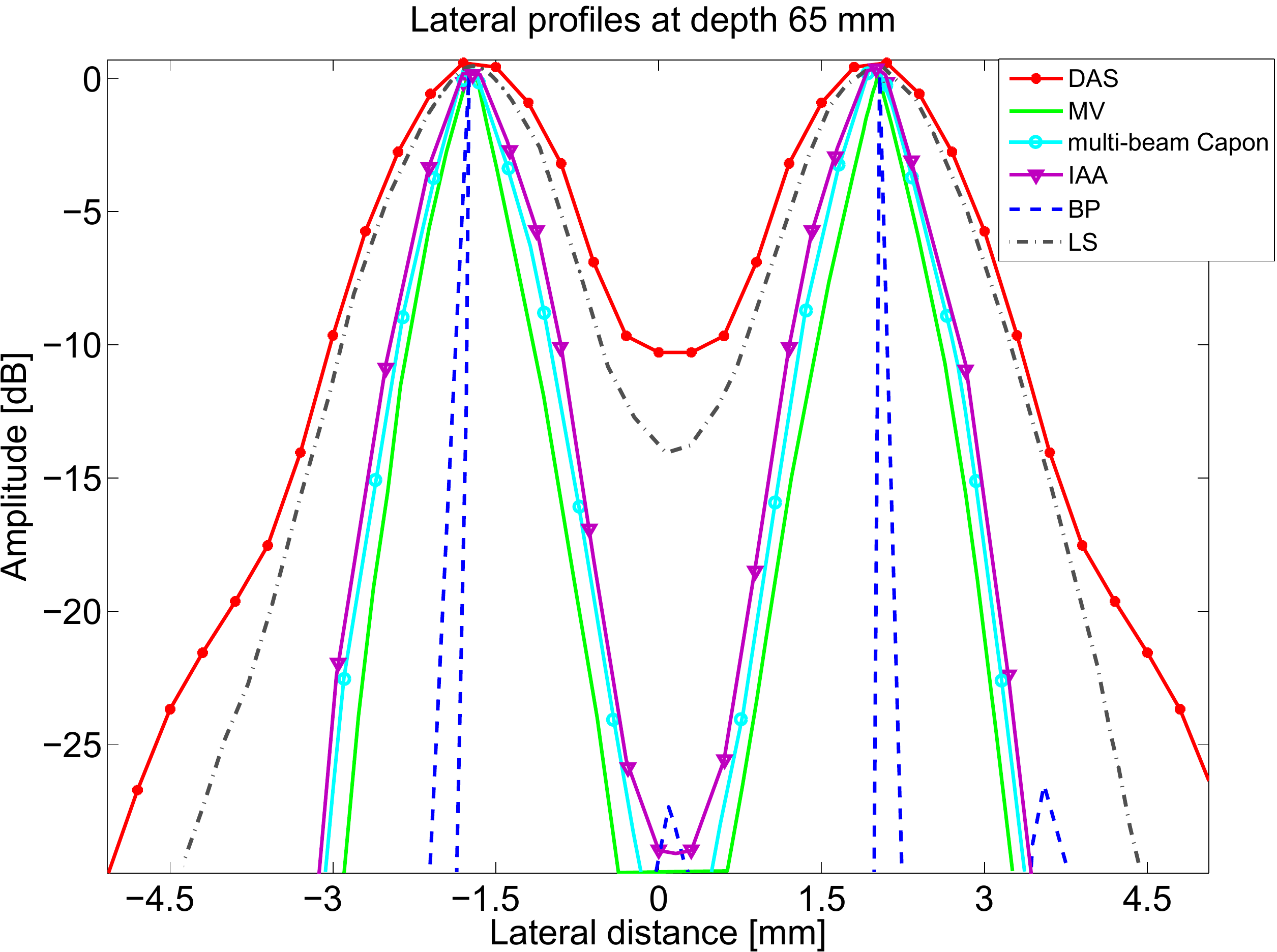}
      \hfil
      \caption{Lateral profiles at 65 mm depth of the point reflectors represented in Fig.~\ref{fig:comp_reflectors}.}
      \label{fig:lat_scatt}
   \end{figure}
   
Fig.~\ref{fig:lat_scatt} presents the lateral profiles of the compared BF methods at 65 mm. We can observe that multi-beam Capon and MV are comparable in terms of lateral profiles, but MV offers better delimitation of the two points. As observed, BP BF outperforms the other BF methods, being able to perfectly resolve the two points, suppressing also the sidelobes. Finally, LS BF gives the smoothest result.
   
\subsection{Simulated hypoechoic cyst}
\begin{figure}
      \center
      \includegraphics[width=0.49\textwidth]{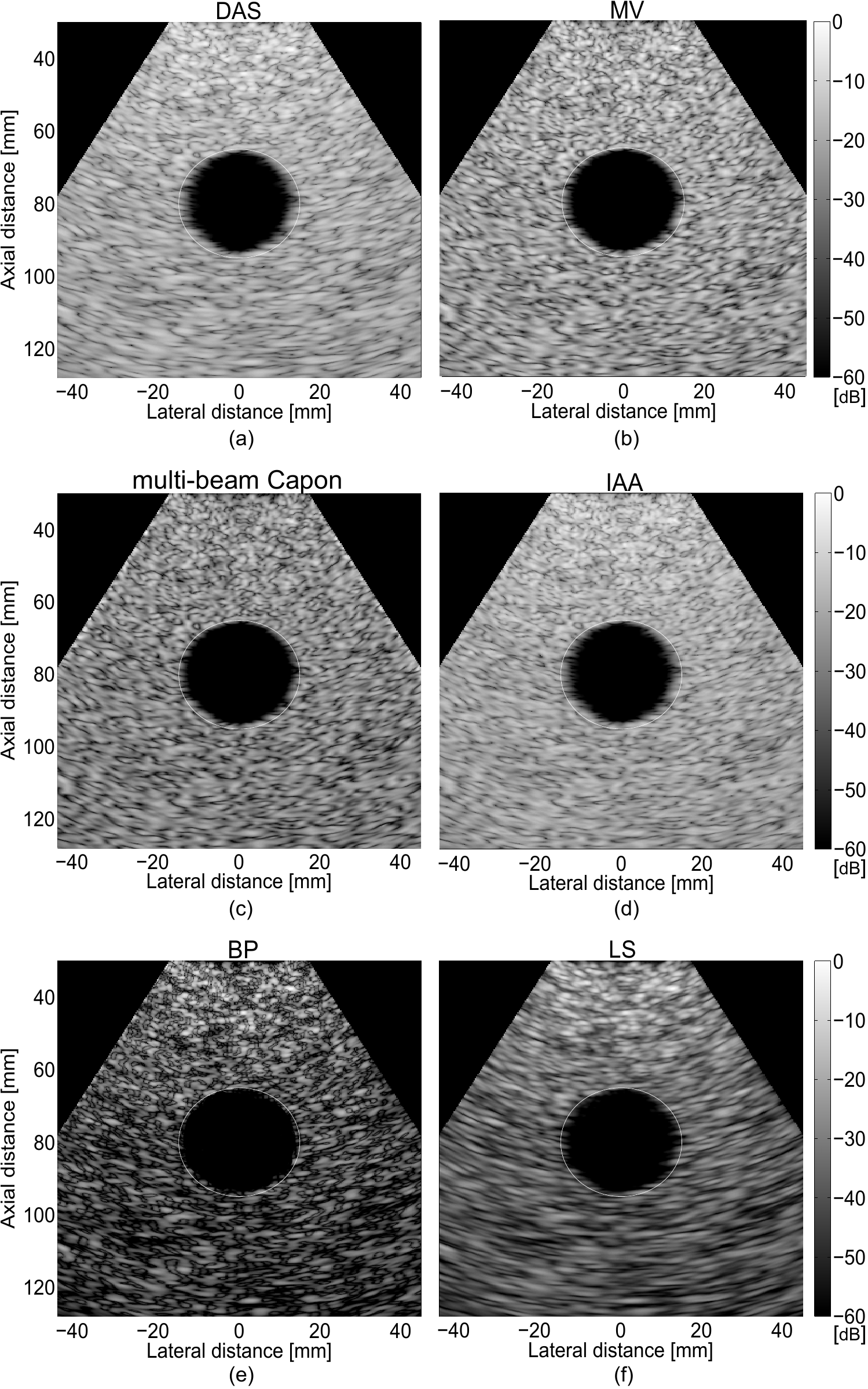}
      \hfil
      \caption{(a) DAS, (b) MV, (c) multi-beam Capon, (d) IAA, (e) BP, and (f) LS BF results of the hypoechoic cyst simulation.}
      \label{fig:comp_hypo_phantom}
   \end{figure}

The BF results of a hypoechoic cyst in a speckle pattern are shown in the Fig.~\ref{fig:comp_hypo_phantom}. We have highlighted with white circle the true borders of the cyst, in order to show the accuracy of the proposed methods regarding the dimensionality of the scanned structures. 
   
\begin{table}
\centering
\caption{CNR, SNR, and RG values for the simulated phantom in Fig.~\ref{fig:comp_hypo_phantom}}
\begin{center}
\begin{tabular}{ l l l l } 
\hline
\noalign{\vskip 2mm}
\textbf{BF Method} & \textbf{CNR} & \textbf{SNR} & \textbf{RG} \\[1ex]
\hline\hline
\noalign{\vskip 2mm}
DAS & 4.8 & 0.4 & 1 \\
\hline
\noalign{\vskip 2mm}
MV & 5.3 & 0.61 & 3.64 \\
\hline
\noalign{\vskip 2mm}
multi-beam Capon & 5.4 & 0.58  & 4.87  \\
\hline
\noalign{\vskip 2mm}
IAA & 3.5 & 0.63 & 3.57  \\
\hline
\noalign{\vskip 2mm}
BP & 6.5 & 0.62 & 8.72 \\
\hline
\noalign{\vskip 2mm}
LS & 7.4 & 0.68 & 2.65 \\
\hline
\end{tabular}
\end{center}
\label{tab:cnr_hypo}
\end{table}

The image quality metrics are detailed in Table~\ref{tab:cnr_hypo}. To calculate the CNR, we have considered the region $R_2$ inside the hypoechoic cyst (the black region), and the region $R_1$ inside the homogeneous speckle, at the same depth and with same dimension as the region $R_2$, as suggested in \cite{rindal_understanding_2014}. The SNR was computed for $R_1$. For calculating RG, the whole image was considered. As expected, with DAS the cyst appears more narrow due to the low resolution and its low capability of resolving cyst-like structures inside the speckle pattern, Fig.~\ref{fig:comp_hypo_phantom}(a). By using MV, we slightly increase the contrast and the resolution in the final image compared with DAS, the dimension of the cyst being closer to its real dimension, as shown in Fig.~\ref{fig:comp_hypo_phantom}(b). Better resolution is obtained when the multi-beam Capon is used, the RG being increased by a factor of almost 1.4. The improve in resolution can be observed also in the delimitation of the cyst region compared with the white circle that represents the real dimension of the cyst, see Fig.~\ref{fig:comp_hypo_phantom}(c). Compared with DAS, IAA increases the resolution of the beamformed image, but not as much as MV or multi-beam Capon, Fig.~\ref{fig:comp_hypo_phantom}(d). However, a contrast degradation can be observed from Table~\ref{tab:cnr_hypo}. Finally, the proposed methods are reflecting more correctly the real dimension of the cyst, especially when using the BP BF, Fig.~\ref{fig:comp_hypo_phantom}(e), this being in concordance with the high increase in resolution (with a factor of 2 compared with MV) and contrast. As expected, LS tends to favor continuity and smoothness, especially when dealing with the speckle pattern (see Fig.~\ref{fig:comp_hypo_phantom}(f)), the gain in resolution being less important. However, even so, it is more precise in reflecting the dimensionality of the cyst. Note that in terms of SNR, in comparison with DAS, all the other beamformers give better SNR, the best improvement being obtained with LS BF which is also outperforming the other beamformers in terms of contrast.     

\begin{figure}
      \center
      \includegraphics[width=0.49\textwidth]{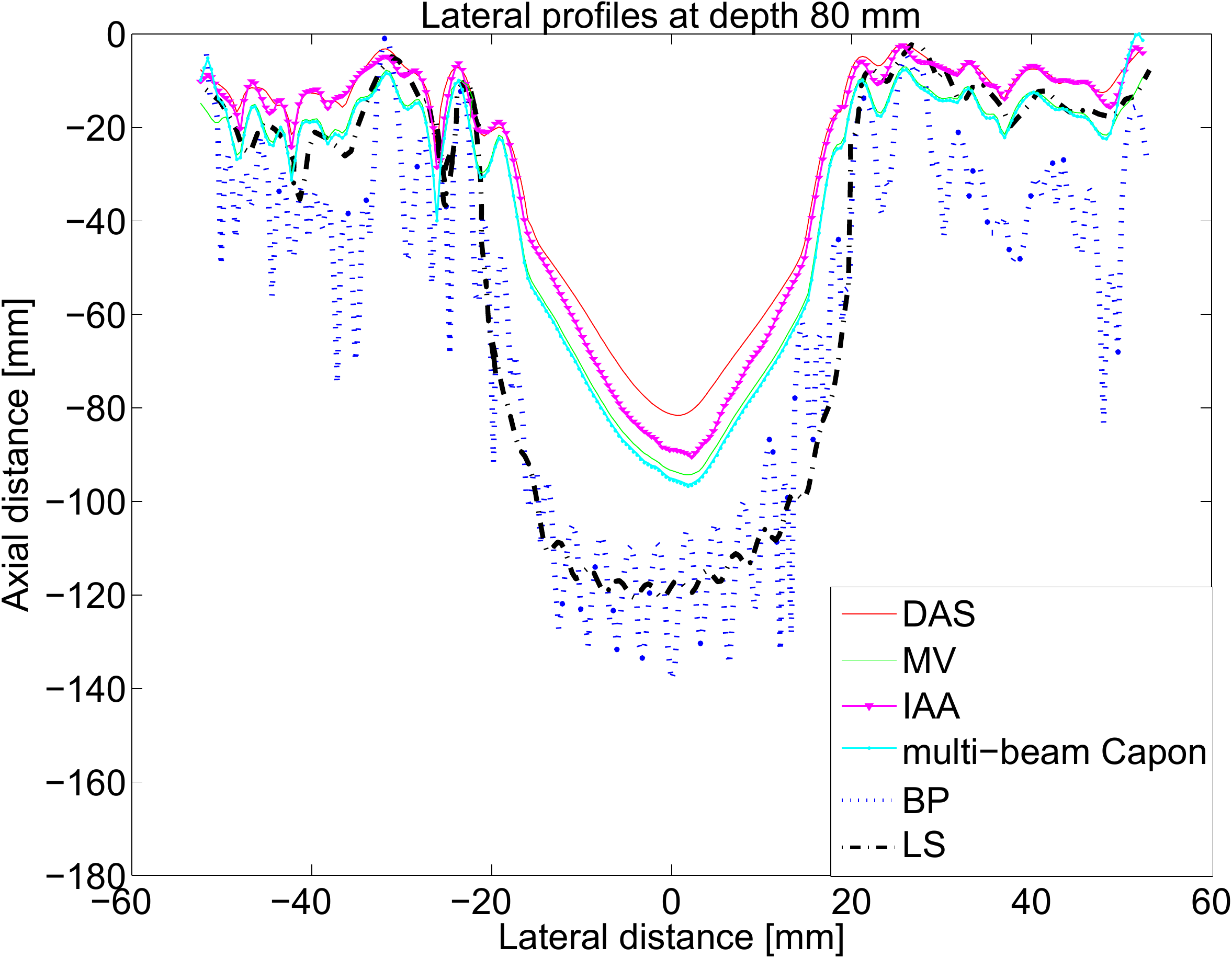}
      \hfil
      \caption{Lateral profiles at 80 mm depth of the cyst phantom represented in Fig.~\ref{fig:comp_reflectors}.}
      \label{fig:lat_hypo_phantom}
   \end{figure}    

Fig.~\ref{fig:lat_hypo_phantom} presents the lateral profiles of the results presented in the Fig.~\ref{fig:comp_hypo_phantom}, where the previous observations are confirmed. The curves in Fig.~\ref{fig:lat_hypo_phantom} are computed by averaging 15 lateral profiles around depth 80 mm. The proposed methods, BP and LS, have larger mainlobes than the other BF methods, that correspond to the true dimension of the hypoechoic cyst. We can also confirm the increase in contrast presented in the Table~\ref{tab:cnr_hypo} in case of BP and LS BF approaches.

\begin{figure}
      \center
      \includegraphics[width=0.49\textwidth]{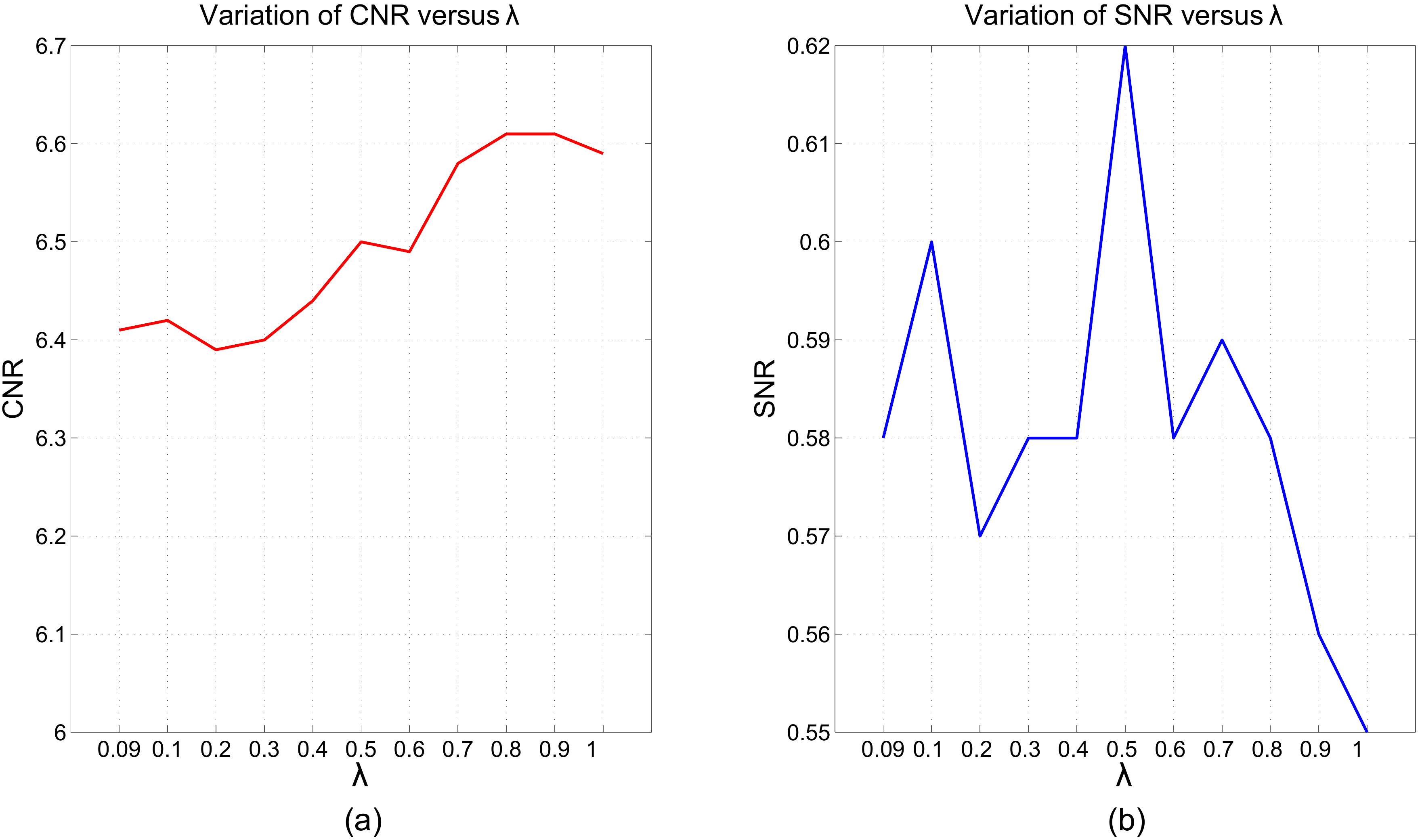}
      \hfil
      \caption{The variation of (a) CNR and (b) SNR versus $\lambda$ when BP method was applied to the hypoechoic cyst simulation.}
      \label{fig:BP_CNR_SNR_variation}
   \end{figure}

Fig.~\ref{fig:BP_CNR_SNR_variation} presents the variation of CNR and SNR parameters function of $\lambda$ hyperparameter. We can observe that a favorable compromise between CNR and SNR is reached when $\lambda=0.5$. The value of CNR can be improved by increasing the value of $\lambda$. For example, when $\lambda=0.9$, CNR$=6.61$, but the value of SNR is reduced, SNR$=0.55$. Similarly, decreasing the value of $\lambda$ will increase the value of SNR, while losing in CNR.

\subsection{Simulated cardiac image} \label{sec:sim_cardiac_res}

\begin{figure}
      \center
      \includegraphics[width=0.49\textwidth]{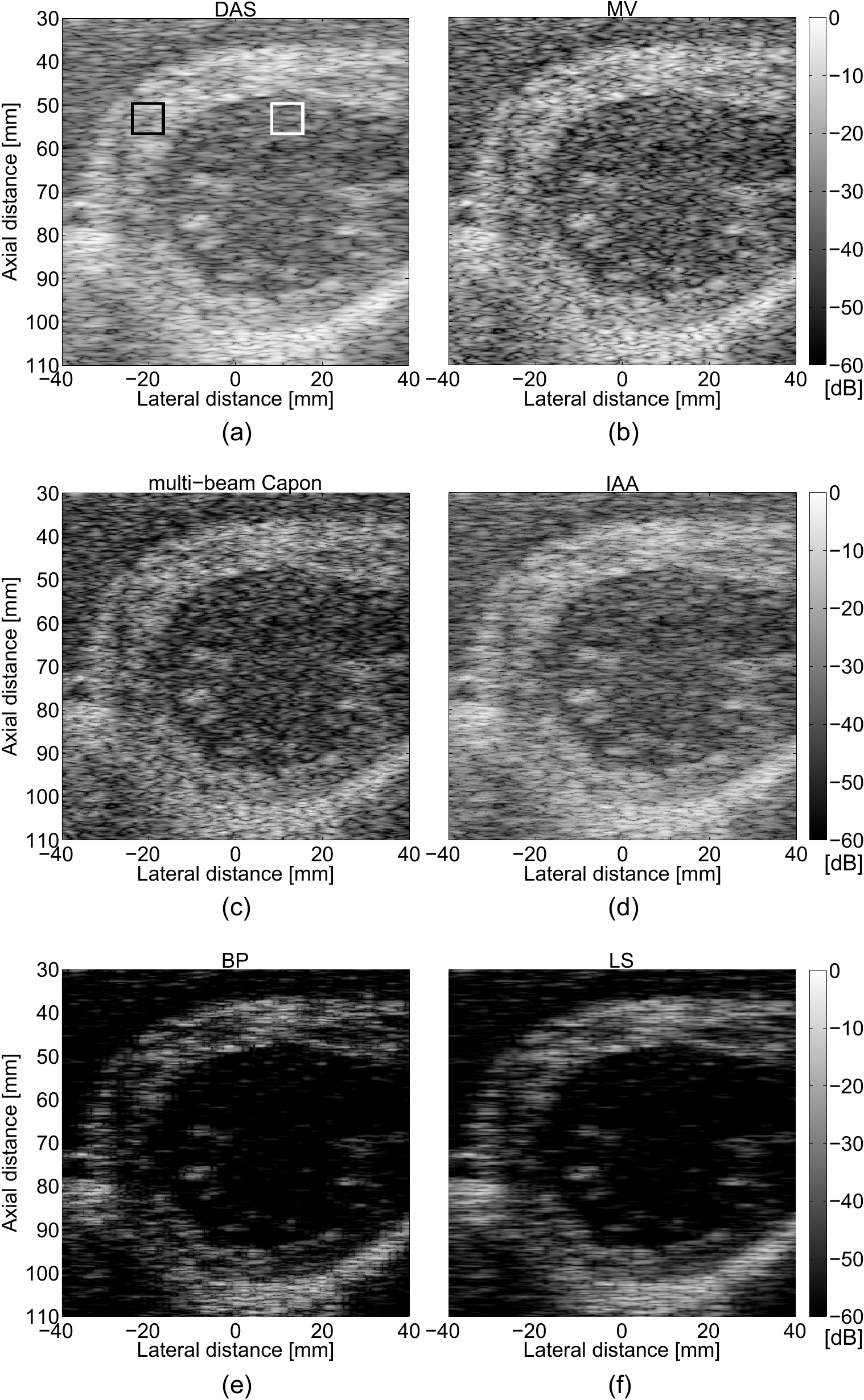}
      \hfil
      \caption{(a) DAS, (b) MV, (c) multi-beam Capon, (d) IAA, (e) BP, and (f) LS BF results of the ultrarealistic simulation of a cardiac image.}
      \label{fig:comp_cardio}
   \end{figure}
   
\begin{table}
\centering
\caption{CNR, SNR, and RG values for the simulated US cardiac beamformed images in Fig.~\ref{fig:comp_cardio} }
\begin{center}
\begin{tabular}{ l l l l} 
\hline
\noalign{\vskip 2mm}
\textbf{BF Method} & \textbf{CNR} & \textbf{SNR} & \textbf{RG} \\[1ex]
\hline\hline
\noalign{\vskip 2mm}
DAS & 1.12 & 0.47 & 1 \\
\hline
\noalign{\vskip 2mm}
MV & 0.90 & 0.56 & 3.56 \\
\hline
\noalign{\vskip 2mm}
multi-beam Capon & 0.61  &  0.54 & 4.23  \\
\hline
\noalign{\vskip 2mm}
IAA & 1.30 & 0.62 & 5.3  \\
\hline
\noalign{\vskip 2mm}
BP & 1.45 & 1.75 &9.89 \\
\hline
\noalign{\vskip 2mm}
LS & 1.55 & 1.88 & 2.06 \\
\hline
\end{tabular}
\end{center}
\label{tab:cnr_cardio}
\end{table}

The results of beamforming the cardiac medium are shown in the Fig.~\ref{fig:comp_cardio}. With this example, we are interested in visualizing the LV region (hypoechoic), which is surrounded by the hyperechoic regions containing the anterior and posterior walls of the heart as well as the septum. The small echoic regions inside the LV region are the papillary muscles, that due to the low contrast and resolution of the DAS beamformed image are hard to be distinguished, Fig.~\ref{fig:comp_cardio}(a). A better visualization of the walls is obtained with MV (Fig.~\ref{fig:comp_cardio}(b)) and multi-beam Capon (Fig.~\ref{fig:comp_cardio}(c)), resulting also in an improved resolution, confirmed with a higher RG value (see Table~\ref{tab:cnr_cardio}). For the calculation of the CNR, we considered $R_1$ the region inside the white square, situated at approximately 18 mm (laterally) and around 55 mm (axially), while $R_2$ is delimited by the black square, around -20 mm (laterally) and 55 mm (axially). To compute SNR, the $R_1$ region was considered.  

An interesting observation is that the value of the CNR in the case of MV and multi-beam Capon is not improved compared with DAS. This is explained by the results in \cite{rindal_understanding_2014}, where it has been shown that the improvement of the contrast directly depends on the high definition of the regions (the LV, the septum, and the walls in our example). Since the amplitude of the reflectors from the walls and septum are not so high compared with the region of LV that contains speckle, the contrast of the final image is affected. However, IAA improve both the contrast and the resolution of the image, presenting more defined regions, as shown in Fig.~\ref{fig:comp_cardio}(d). Yet, the best improvement of the resolution is obtained when we promote Laplacian BF solutions, with BP BF, see Fig.~\ref{fig:comp_cardio}(e), resulting in an improvement by a factor of 2 in RG compared with MV and multi-beam Capon, and by a factor of almost 10 compared with DAS. Of course, as expected, LS BF is highly improving the contrast and the SNR of the resulted image, while the RG is lower than when using the other BF approaches, Fig.~\ref{fig:comp_cardio}(f).

\subsection{\textit{In vivo} data: carotid}
\begin{figure}
      \center
      \includegraphics[width=0.49\textwidth]{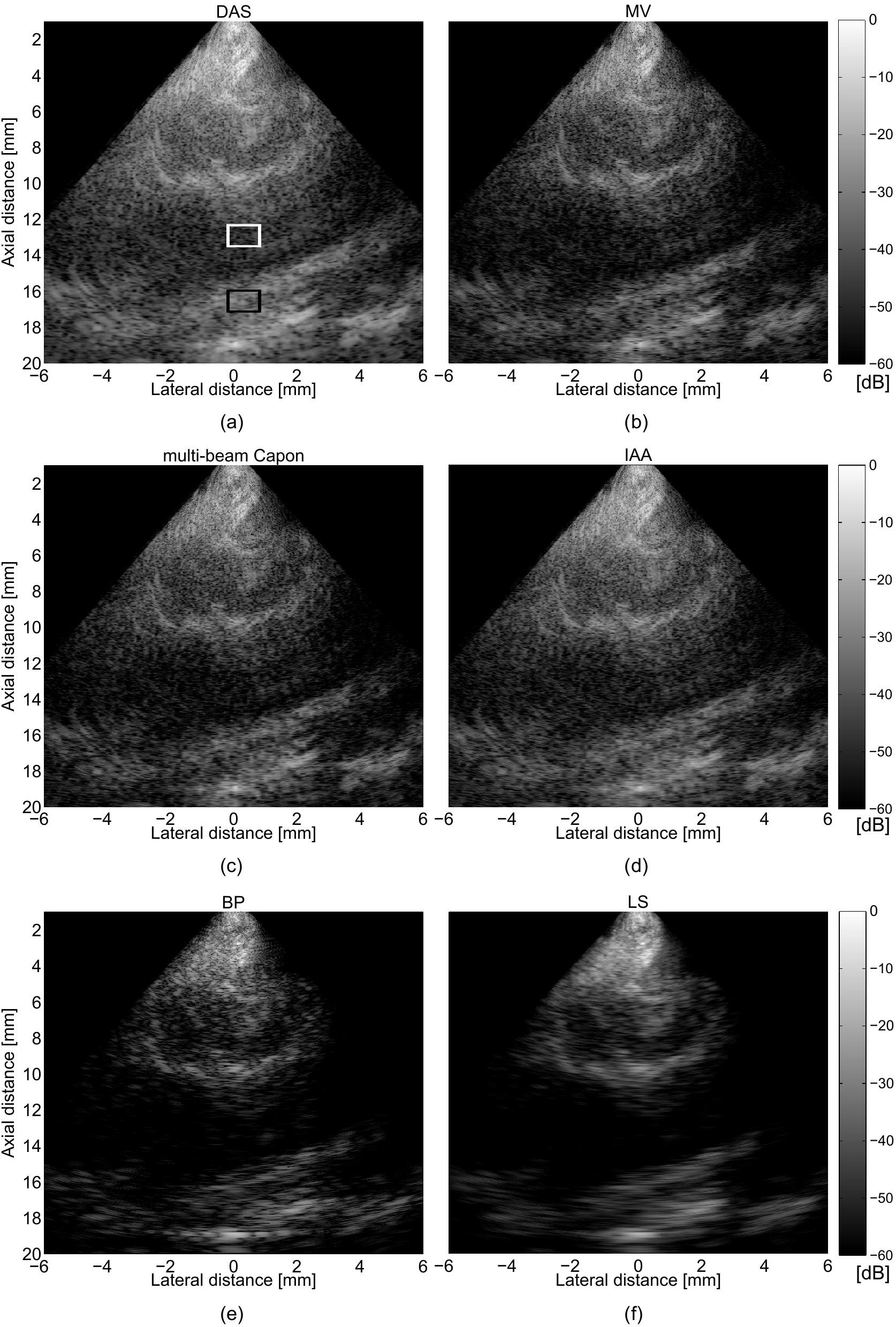}
      \hfil
      \caption{(a) DAS, (b) MV, (c) multi-beam Capon, (d) IAA, (e) BP, and (f) LS BF results of experimental carotid data.}
      \label{fig:comp_carotid}
   \end{figure}
   
\begin{table}
\centering
\caption{CNR, SNR, RG, and computational time values for the experimental carotid beamformed images from Fig.~\ref{fig:comp_carotid} }
\begin{center}
\begin{tabular}{ l l l l | c } 
\hline
\noalign{\vskip 2mm}
\textbf{BF Method} & \textbf{CNR} & \textbf{SNR} & \textbf{RG} & Computation time [s] \\[1ex]
\hline\hline
\noalign{\vskip 2mm}
DAS & 1.84 & 1.46 & 1 & 4.5552 \\
\hline
\noalign{\vskip 2mm}
MV & 2.24 & 1.43 & 1.25 & 122 \\
\hline
\noalign{\vskip 2mm}
multi-beam Capon & 1.32 & 1.49  & 1.25 & 368 \\
\hline
\noalign{\vskip 2mm}
IAA & 1.48 & 1.47 & 1.34 & 8.9266 \\
\hline
\noalign{\vskip 2mm}
BP & 1.85 & 1.49 & 1.48 & 60.4320 \\
\hline
\noalign{\vskip 2mm}
LS & 1.94 & 1.55 & 1.17 & 8.8692 \\
\hline
\end{tabular}
\end{center}
\label{tab:cnr_caro}
\end{table}
   
Fig.~\ref{fig:comp_carotid} presents the BF results of the studied beamformers, and in the Table~\ref{tab:cnr_caro} we calculated their corresponding CNR, SNR, RG, and computational time values. In this example, the carotid is placed between 8 and 15 mm in the axial direction. In this region, the interior of the carotid artery is the hypo-echoic structure surrounded by the arterial walls (which are hyper-echoic). To calculate the CNR, we have considered region $R_2$ inside the carotide (the white rectangle positioned at $0$ mm laterally), and the region $R_1$ inside the region of speckle (the black rectangle positioned at $0$ mm laterally). The SNR for $R_1$ was computed.

As observed, by using DAS BF is hard to distinguish between the interior of the carotid and its walls, Fig.~\ref{fig:comp_carotid}(a). This can be also explained by the fact that DAS BF result represents the lower RG. A better visualization of the structures of interest are obtained with MV and multi-beam Capon, that have similar RG values. However, the contrast of the MV beamformed image is better, increasing the value of CNR by a factor of $\approx 1$ compared to multi-beam Capon. We can observe that multi-beam Capon is clearly defining the region inside the carotid, by reducing the level of speckle inside it, cf. Fig.~\ref{fig:comp_carotid}(c). The IAA beamformed image is comparable with the one of multi-beam Capon, but it conserves better the speckle inside the carotid, offering a better resolution and a better contrast of the image. With the proposed approaches, however we are able to better distinguish the interior of the carotid artery, as well as its walls, with a high gain in contrast and resolution resulted by applying BP BF. A loss in resolution can be observed when using LS, compared with BP, due to the use of the $\ell_{2}$-norm regularization. Note that, due to the formulation of the proposed direct model \eqref{eq:model_das_not_beamspaced} that includes an additive noise, the proposed method is intrinsically denoising the signal (e.g., the noise inside the carotid is reduced) through the inversion process (see Table~\ref{tab:cnr_caro}). The denoising effect obtained by our BF approach does not suffer from any spatial resolution loss, as it could be the case if the raw data or the beamformed images were low-pass filtered.

Regarding the computational time, note that it is highly dependent on the length of the acquired raw data. For this case, the number of ranges was around $N=2500$, and the proposed approaches were applied without a previous decimation of the raw data. Of course, the standard parallel computing methods could additionally improve the computational complexity, since the BF process is done for each lateral scanline. All the discussed methods were implemented with Matlab R2013b, on an Intel i7 2600 CPU working at 3.40GHz. Note that even so, LS BF is approaching the time capabilities of DAS, being just twice slower than DAS. Moreover, BP is also faster than MV. Thus, by using the discussed techniques for improving the computational expense, makes the two proposed methods good candidates for real-time applications.

\subsection{\textit{In vivo} data: healthy thyroid}
\begin{figure}
      \center
      \includegraphics[width=0.51\textwidth]{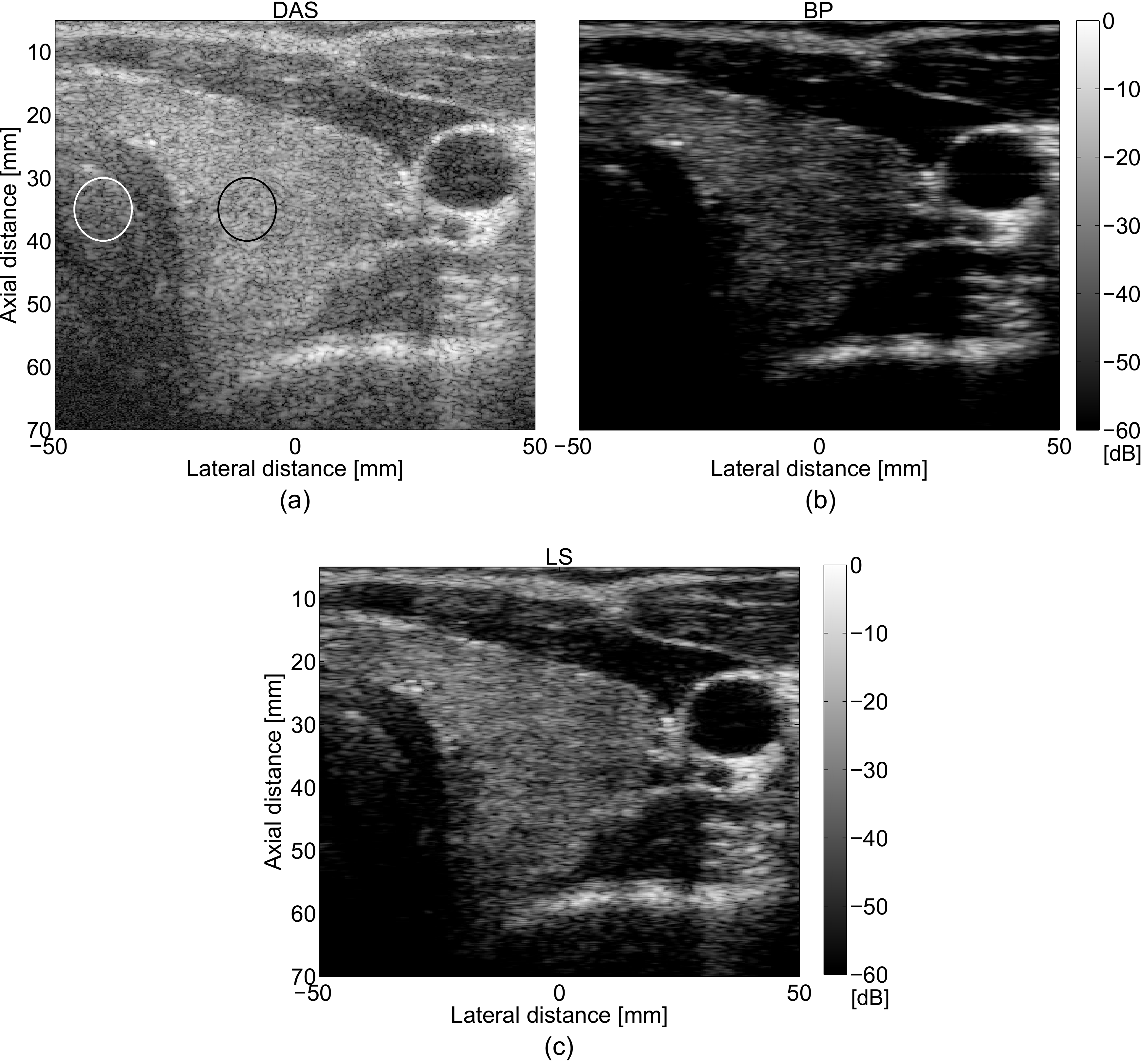}
      \hfil
      \caption{(a) DAS, (b) BP, and (c) LS BF results of \textit{in vivo} helthy thyroid data.}
      \label{fig:comp_thyroid_healthy}
   \end{figure}
   
Fig.~\ref{fig:comp_thyroid_healthy} presents the beamforming results of healthy thyroid data. The thyroid (echoic region) is situated between the trachea and the carotid artery (laterally, between -20 mm and 30 mm approximately). Fig.~\ref{fig:comp_thyroid_healthy} (a) illustrates the result obtained with DAS BF. As expected, the contrast of the image is low and it is hard to distinguish the thyroid structure from the trachea, especially in the upper-left part of the thyroid. However, when BP (Fig.~\ref{fig:comp_thyroid_healthy} (b)) and LS Fig.~\ref{fig:comp_thyroid_healthy} (c) are used, the thyroid region is easy to be identified, and the contrast of the image is increased.
\begin{table}
\centering
\caption{CNR, SNR, and RG values for the \textit{in vivo} healthy thyroidal beamformed images from Fig.~\ref{fig:comp_thyroid_healthy}}
\begin{center}
\begin{tabular}{ l l l l} 
\hline
\noalign{\vskip 2mm}
\textbf{BF Method} & \textbf{CNR} & \textbf{SNR} & \textbf{RG} \\[1ex]
\hline\hline
\noalign{\vskip 2mm}
DAS & 0.55 & 0.22 & 1 \\
\hline
\noalign{\vskip 2mm}
BP & 1.13 & 0.32 & 3 \\
\hline
\noalign{\vskip 2mm}
LS & 1.56  & 0.64 & 2.5 \\
\hline
\end{tabular}
\end{center}
\label{tab:cnr_thyro_healthy}
\end{table}

The values of CNR, SNR, and RG are depicted in the Table~\ref{tab:cnr_thyro_healthy}. To compute CNR we considered region $R_2$ inside the thyroid (the black circle positioned at approximately $-10$ mm laterally), and the region $R_1$ inside trachea (the white circle positioned at approximately $-40$ mm laterally). The SNR for $R_1$ was computed. We can observe that the best values of the CNR and SNR are obtained when LS method was applied, the thyroid region being obvious to be discerned. The boundaries of the carotid artery are also well defined, Fig.~\ref{fig:comp_thyroid_healthy} (c).

\subsection{\textit{In vivo} data: thyroid with tumor}
\begin{figure}
      \center
      \includegraphics[width=0.49\textwidth]{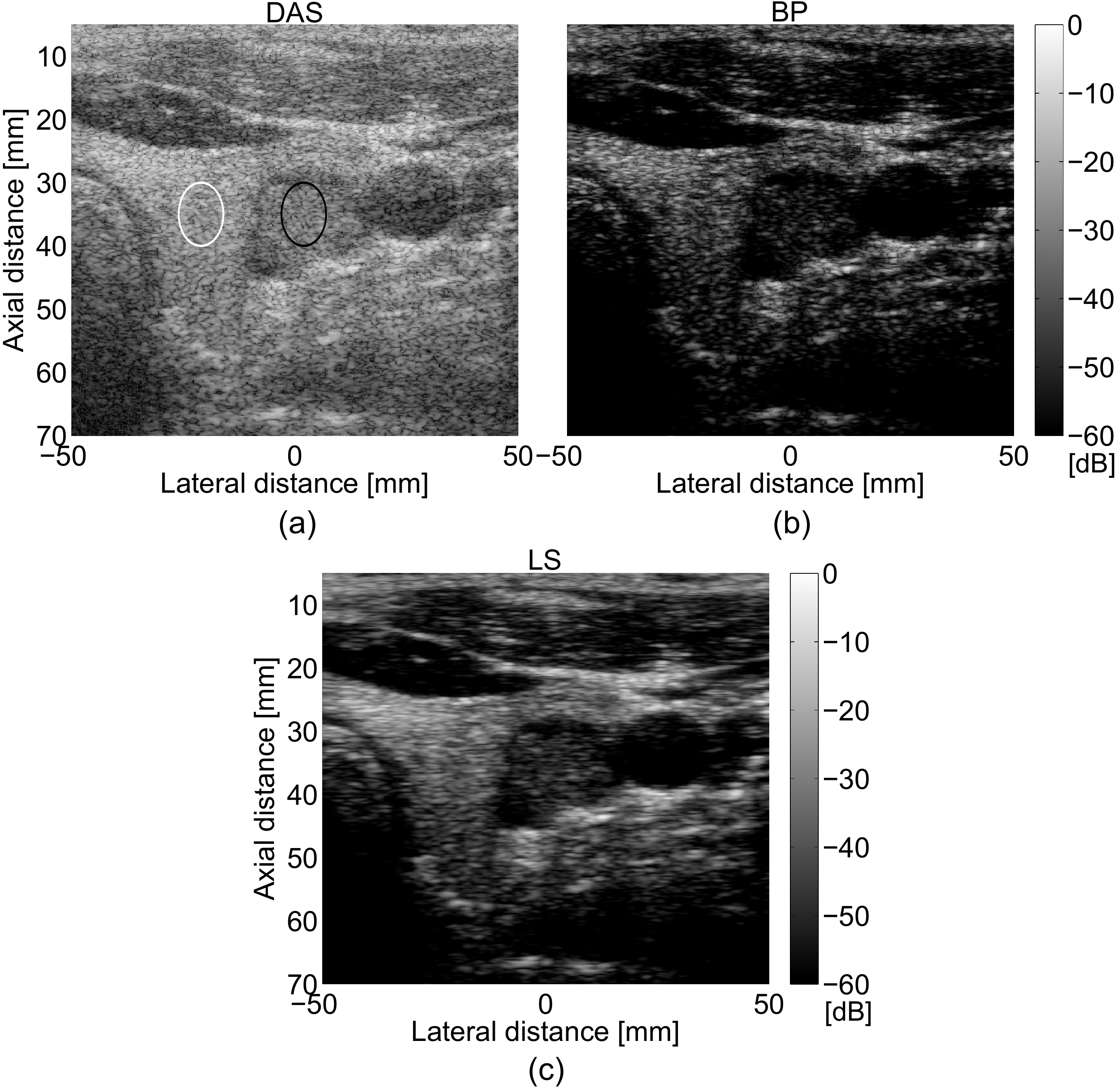}
      \hfil
      \caption{(a) DAS, (b) BP, and (c) LS BF results of \textit{in vivo} thyroid data with tumor.}
      \label{fig:comp_thyroid}
   \end{figure}
   
\begin{table}
\centering
\caption{CNR, SNR, and RG values for the \textit{in vivo} thyroidal beamformed images from Fig.~\ref{fig:comp_thyroid}}
\begin{center}
\begin{tabular}{ l l l l} 
\hline
\noalign{\vskip 2mm}
\textbf{BF Method} & \textbf{CNR} & \textbf{SNR} & \textbf{RG} \\[1ex]
\hline\hline
\noalign{\vskip 2mm}
DAS & 0.71 & 0.62 & 1 \\
\hline
\noalign{\vskip 2mm}
BP & 1.16 & 0.79 & 2.9 \\
\hline
\noalign{\vskip 2mm}
LS & 1.32  & 0.86 & 1.5 \\
\hline
\end{tabular}
\end{center}
\label{tab:cnr_thyro}
\end{table}

The beamformed results of the thyroid data with tumor are presented in the Fig.~\ref{fig:comp_thyroid}. The malignant tumor with an irregular structure can be seen between the left lobe of the thyroid (the hyper-echoic structure situated near the trachea) and the carotid artery (the hypo-echoic circular structure with the center at approximately 33 mm (axially) and 40 mm (laterally). We can observe that, contrarily to DAS beamformed image, where the tumor is hard to be distinguished (see Fig.~\ref{fig:comp_thyroid} (a)), both our methods improve the visualization of the main structures, enhancing the edges of the tumor. The values of CNR, SNR, and RG are depicted in the Table~\ref{tab:cnr_thyro}, where a gain in resolution with a factor of almost 3 can be observed when using BP, compared with DAS, while with LS we obtain a higher improvement in contrast and SNR than with BP BF. CNR was computed by considering region $R_2$ inside the tumor (the black circle positioned at $0$ mm laterally), and the region $R_1$ inside the left lobe of the thyroid (the white circle positioned at approximately $-20$ mm laterally). The SNR for $R_1$ was computed.

\section{Conclusion} \label{sec:conclusion}
We have presented a new BF approach in US medical imaging, that solves a regularized inverse problem based on a linear model relating, for each depth, the US reflected data to the signal of interest. Contrarily to existing techniques that use adaptive or non-adaptive weights to average the raw data in order to form RF lines, we directly recover, for each depth, the desired signals using Laplacian or Gaussian statistical assumptions. The proposed regularization-based BF allows us to take advantage of the beamspace processing that enables to highly reduce the number of US transmissions (by a factor of five in our examples), while improving the quality of the beamformed images compared to four existing beamformers. Multiple simulated and experimental examples were presented, that compare our approach with DAS, MV, multi-beam Capon, and IAA beamformers. We showed that our BF approaches, based on Laplacian and Gaussian prior information, although based on the same model, are complementary in terms of result quality. Thus, Laplacian statistics are favoring sparse results while the Gaussian law is offering more regular and smooth images. We also proved through resolution gain, CNR and SNR image quality metrics, that we obtained an important gain in spatial resolution and/or in contrast, while maintaining a reasonable computational time compared to other existing techniques.
As future work, we will consider other statistical assumptions, such as the generalized Gaussian distribution, resulting in $\ell_{p}$-norm minimization with the parameter $p$ between $0$ and $2$. Following the choice of $p$, this should guarantee a better compromise between the gain in contrast and the improve of the spatial resolution (e.g. \cite{alessandrini_restoration_2011}, \cite{zhao_restoration_2014}, and \cite{ChBaKo2016.1}). Another way to obtain this compromise could be to combine the two regularization terms (through Laplacian and Gaussian statistics) used in our approach, resulting in an elastic net regularization (see e.g., \cite{zou_regularization_2005} and \cite{Szasz2016}).

Another interesting perspective offered by our BF direct linear model is the possibility to combine it with existing post-processing techniques, aiming to enhance the quality of US images, such as deconvolution or super-resolution.

\appendices
\section{Minimum variance beamforming} \label{subsec:mv}

MV (or Capon filter) BF \cite{capon1969high} consists in minimizing the array output power by maintaining a unit gain at the focal point. It adaptively calculates the weights, by solving:

\begin{equation}
\min_{\boldsymbol{w}}\boldsymbol{w}^{H}\boldsymbol{R}_{k}\boldsymbol{w}, \ \ \ \  \text{such that} \ \ \ \ \boldsymbol{w}^{H}\boldsymbol{1}=1, 
\label{eq:mv_crit}
\end{equation}
with the analytical solution: 
\begin{equation}
\boldsymbol{w}_{MV}=\frac{\boldsymbol{R}_{k}^{-1}\boldsymbol{1}}{\boldsymbol{1}^{T}\boldsymbol{R}_{k}^{-1}\boldsymbol{1}},
\label{eq:mv_w}
\end{equation}
where $\boldsymbol{R}_{k}=E[\boldsymbol{y}_{k}\boldsymbol{y}_{k}^{H}]$ is the covariance matrix of $\boldsymbol{y}_{k}$ and $\boldsymbol{1}$ is a length $M$ column-vector of ones. These weights are then used to calculate the desired RF beamformed lines in a similar way as with DAS.  
In practice, $\boldsymbol{R}_{k}$ is unavailable and the estimated covariance matrix $\hat{\boldsymbol{R}}_{k}$ is used as alternative, derived from $L$ received samples:

\begin{equation}
\hat{\boldsymbol{R}}_{k}=\sum_{l=1}^{L}\boldsymbol{y}_{k}(l)\boldsymbol{y}_{k}^{H}(l).
\label{eq:cov_mat}
\end{equation}

Since the received focused raw data is coherent, several methods were proposed to decorrelate the data as much as possible: subaperture (or subarray) averaging (also called spatial smoothing), time averaging, and diagonal loading significantly improve the standard MV BF (e.g., \cite{synnevag_benefits_2009}).

\section{Beamspace beamforming} \label{subsec:beamsp_beam_Butler}

Starting from the MV BF method presented in Section \ref{subsec:mv}, named element-space based Capon (ES-Capon) in \cite{nilsen_beamspace_2009}, Nilsen \textit{et. al.} proposed a beamspace beamformer (BS-Capon) that allowed reducing the computational complexity of the MV BF by a ratio of 3. Basically, they reduce the size of the covariance matrix in \eqref{eq:cov_mat} by replacing it with a smaller covariance matrix of orthogonal beams. The expression of the orthogonal beams is detailed in \cite{nilsen_beamspace_2009}, and the beamspace transformation is expressed as follows:

\begin{equation}
\boldsymbol{y}_{k_{BS}} = \boldsymbol{B}\boldsymbol{y}_{k}, 
\label{eq:beamspace_transf}
\end{equation}
where $\boldsymbol{B}=[\boldsymbol{b}_{1},\cdots,\boldsymbol{b}_{M}]^T$ is the $M \times M$ Butler matrix whose elements are defined as:

\begin{equation}
b_{mn}=\frac{1}{\sqrt{M}}e^{j\frac{2\pi}{M}(m-\frac{1}{2})n}.
\label{eq:butler}
\end{equation}
$\boldsymbol{B}$ is an unitary matrix ($\boldsymbol{B}\boldsymbol{B}^{H}=\boldsymbol{B}^{H}\boldsymbol{B}=\boldsymbol{I}_{M \times M}$), equivalent to an M-point discrete Fourier transform (DFT) matrix. $\boldsymbol{I}_{M \times M}$ is the identity matrix of size $M \times M$.

The transformation in \eqref{eq:beamspace_transf} is applied to all signals and weights vectors in the element-space (ES) to find their beamspaced version. Therefore, the weights of ES-Capon BF are formed by solving:

\begin{equation}
\min_{\boldsymbol{w}_{BS}}\boldsymbol{w}_{BS}^{H}\boldsymbol{R}_{BS}\boldsymbol{w}_{BS}, \ \ \ \  \text{such that} \ \ \ \ \boldsymbol{w}_{BS}^{H}e_{1}=1.
\label{eq:mv_crit_bs}
\end{equation}
The solution of \eqref{eq:mv_crit_bs} is:

\begin{equation}
\boldsymbol{w}_{BS}=\frac{\boldsymbol{R}_{BS}^{-1}e_{1}}{e_{1}^{T}\boldsymbol{R}_{BS}^{-1}e_{1}},
\label{eq:mv_w_bs}
\end{equation}
where $\boldsymbol{R}_{BS}=E[\boldsymbol{y}_{k_{BS}}\boldsymbol{y}_{k_{BS}}^{H}]$ is the covariance matrix of $\boldsymbol{y}_{k_{BS}}$ and $e_{m}$ is a $M \times 1$ vector having the value $1$ in the $m$-th position and zero in all other positions. Finally, we can state that BS-Capon BF can be seen as the description of the Capon filter from \eqref{eq:mv_w} in the Fourier domain. 

As stated before in this section, in the case of DAS, MV, and BS-Capon BF, the final RF US image is a collection of RF beamformed lines, each of which being the result of beamforming the raw RF signals coming from a focused wave emission in the direction $\theta_{k}$, $k\in \{1,\hdots,K\}$, using the $M$ elements of the transducer.

The most commonly used visualization mode in ultrasound medical imaging is the B (brightness)-mode. It is obtained by applying envelope detection and log-compression techniques to each beamformed RF line. Finally all the RF lines are juxtaposed in the lateral direction to form the final 2D US image, as shown in the Fig.~\ref{fig:setup}.

\section{Multi-beam Capon beamforming} \label{subsec:multibean_caponBF}

Jensen \textit{et. al.} used beamspace processing for reducing the dimensionality of the data, and proposed a new approach of Capon BF, called multi-beam Capon BF \cite{jensen_approach_2012}. For more convenience, let us briefly recall their approach.

For a given range $n$, let us select its corresponding lateral scanline, as illustrated in Fig.~\ref{fig:setup}. Since the signals $\boldsymbol{y}_{k_{BS}}$ have been focused in axial direction (by applying time delays) before being beamformed, we just need to compensate the phase-shifts based on the distances from the samples of the lateral scanline (equivalent, in our case, with $K$, the number of beam directions). 

\begin{figure}
  \centering
  \includegraphics[width=0.40\textwidth]{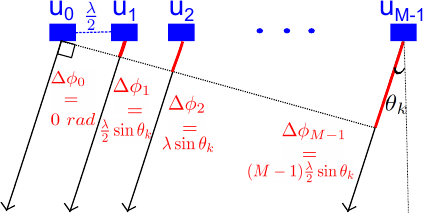}
  \caption{Phase shift compensation of the focused raw data.}
\label{fig:phase_shift}
\end{figure}

The compensation of the phase-shifts, $\Delta \phi _{m}$, with $M=0, \cdots, M-1$ is depicted in Fig.~\ref{fig:phase_shift}, assuming that the time-compensated data reaches the elements at angle $\theta_{k}$. We consider the first elements as reference, so its phase-shift is $0$. Thus, based on the far-field assumption, we can formulate the complex exponential version of the manifold vector for a given direction $k$, that corresponds to the incident angle $\theta _{k}$, as (see e.g. \cite{jensen_approach_2012} and \cite{kautz_beamspace_1996}):

\begin{equation}
\boldsymbol{a}_{\theta _{k}}=[1 \ \  e^{-j \pi \sin(\theta _{k})} \cdots e^{-j (M-1) \pi \sin(\theta _{k})}]^{T}.
\label{eq:theta_k_iaa}
\end{equation}

Thus, by using phase shifts, for focusing along a lateral scanline, contrarily to the matrix $\boldsymbol{R}_{BS}$ used in BS-Capon, for a given range $n$, the covariance matrix $\boldsymbol{R}[n]$ will cover all the directions $\theta_{k}, k=1, \cdots, K$. Therefore, the weights corresponding to a given direction $\theta$ and a range $n$ will be formed by solving:

\begin{equation}
\min_{\boldsymbol{w}}\boldsymbol{w}^{H}\boldsymbol{R}[n]\boldsymbol{w}, \ \ \ \  \text{such that} \ \ \ \ \boldsymbol{w}^{H}\boldsymbol{a}_{\theta,n}=1, 
\label{eq:mv_crit_lat}
\end{equation}
having the solution: 
\begin{equation}
\boldsymbol{w}_{\theta,n}=\frac{\boldsymbol{R}^{-1}[n]\boldsymbol{a}_{\theta,n}}{\boldsymbol{a}_{\theta,n}^{T}\boldsymbol{R}^{-1}[n]\boldsymbol{a}_{\theta,n}}.
\label{eq:mv_w_lat}
\end{equation}
These weights are then applied to calculate the signal corresponding to a lateral scanline, at a range $n$.

\section{Iterative adaptive approach beamforming}
\label{subsec:iaa}
Based on the beamspace processing technique and on the calculation of the multibeam covariance matrix discussed in the Section \ref{subsec:multibean_caponBF}, Jensen \textit{et al.} applied IAA \cite{yardibi_nonparametric_2008} to US medical imaging, \cite{jensen_iterative_2014}. Following this recent work, a covariance matrix, $\bar{\boldsymbol{R}}[n]$ based on $\bar{K}$ potential reflectors placed across a considered lateral scanline, was defined as:

\begin{equation}
\bar{\boldsymbol{R}} [n]= \sum_{k=1}^{\bar{K}}|\boldsymbol{y}_{BS}[n]|^{2}\boldsymbol{a}_{\theta}\boldsymbol{a}_{\theta}^{T}=\boldsymbol{A}_{BS}\boldsymbol{P}\boldsymbol{A}_{BS}^{T},
\label{eq:r_modeled}
\end{equation}
with $\boldsymbol{y}_{BS}[n] \in \mathbb{C}^{N_{b}\times \bar{K}}$ the beamspaced time-delayed raw data at a given range $n$, before applying the phase-shift transform. $\boldsymbol{A}$ is the matrix containing the manifold column-vectors defined in \eqref{eq:theta_k_iaa}, and $\boldsymbol{P}$ a diagonal matrix with the elements of $|\boldsymbol{y}_{BS}[n]|^{2}$ along its diagonal. The values of $\boldsymbol{P}$ are then iteratively updated and calculated by taking into account the weights corresponding to a lateral scanline, by following \eqref{eq:mv_w_lat}. Finally, $\boldsymbol{P}$ is used to estimate the amplitude of each reflector of the IAA BF result.   

Contrarily of DAS, MV and BS-Capon BF where, to form the final beamformed image, the RF lines are juxtaposed in the lateral direction, multi-beam Capon and IAA BF are axially juxtaposing the beamformed lateral scanlines to form the final beamformed image.

\section*{Acknowledgement}
We wish to thank Adeline Bernard and Herv\'e Liebgott, from CREATIS laboratory, University of Lyon, for providing the experimental ultrasound data. We are also thankful to Are Charles Jensen for proving us the IAA BF Matlab code and to Dmitry M. Malioutov for the different source localization code examples. This work was partially supported by ANR-11-LABX-0040-CIMI within the program ANR-11-IDEX-0002-02 of the University of Toulouse.  

\bibliographystyle{IEEEtran}	
\footnotesize
\bibliography{Biblio_BF_inverse_pb}

\end{document}